\documentclass[a4paper,11pt]{article}

\usepackage[margin=1in]{geometry} 

\usepackage{amsmath}
\usepackage{amsthm}
\usepackage{amssymb}
\usepackage{lineno}
\usepackage{subcaption}  

\usepackage{dutchcal}

\usepackage[authoryear]{natbib}

\usepackage{placeins}

\usepackage{etoolbox}

\usepackage{graphicx}
\usepackage{bm}
\usepackage{lineno}
\usepackage{caption}

\usepackage{booktabs}

\usepackage[breaklinks]{hyperref}

\usepackage{multirow}

\usepackage{algorithm}
\usepackage{algpseudocode}
\usepackage{bm}
\usepackage{subcaption}
\usepackage{tabularx} 
\usepackage{booktabs}  
\usepackage{float}
\usepackage{placeins}

\usepackage{comment}
\usepackage{xargs}                      

\hypersetup{
	unicode,
	colorlinks=true,
	linkcolor=blue,
	filecolor=magenta,      
	citecolor=blue,
	urlcolor=cyan,
	pdfauthor={ M. A. Mendez},
	pdftitle={Linear and Nonlinear Dimensionality Reduction from Fluid Mechanics to Machine Learning},
	pdfsubject={Linear and Nonlinear Dimensionality Reduction from Fluid Mechanics to Machine Learning},
	pdfkeywords={article, template, simple},
	pdfproducer={LaTeX},
	pdfcreator={pdflatex}
	colorlinks=true,
	citecolor=blue,
	linkcolor=blue,
	filecolor=magenta,      
	urlcolor=blue,
}

\usepackage[most]{tcolorbox}
\usepackage{booktabs}
\usepackage{amsmath}
\usepackage{wrapfig}

\usepackage{mathtools}
\tcbset{enhanced,colback=cyan!2!white,colframe=cyan!90!white,coltitle=blue!20!black,fonttitle=\bfseries}

\usepackage{hyperref}
\hypersetup{
	colorlinks=true,
	linkcolor=blue,
	filecolor=magenta,      
	citecolor=blue,
	urlcolor=cyan,
}

\usepackage{mathrsfs} 

\usepackage{siunitx}
\newcolumntype{Z}[1]{S[scientific-notation=fixed, fixed-exponent=#1, table-format=1.2]}

\title{\LARGE \textbf{Reinforcement Twinning for Hybrid Control of Flapping-Wing Drones}}

\author{
  Romain Poletti$^{1}$\thanks{Corresponding author: romain.poletti@vki.ac.be} \and
  Lorenzo Schena$^{2}$ \and
  Lilla Koloszar$^{1}$ \and
  Joris Degroote$^{3}$  \and
  Miguel Alfonso Mendez$^{1,3,4}$
}

\date{%
\begingroup
\small
\textit{%
$^1$ Environmental and Applied Fluid Dynamics, von Karman Institute for Fluid Dynamics, Belgium \\
$^2$ Department of Mechanical Engineering, Vrije Universiteit Brussel, Belgium \\
$^3$ Department of Electromechanical, Systems and Metal Engineering, Ghent University, Belgium \\
$^3$ Aero-Thermo-Mechanics Laboratory, École Polytechnique de Bruxelles, Université Libre de Bruxelles, Belgium} \\
$^4$ Experimental Aerodynamics and Propulsion Lab, Universidad Carlos III de Madrid, Spain \\
\endgroup
}

\begin{document}
\maketitle
	
\begin{abstract}

Controlling flapping-wing drones requires controllers that handle time-varying, nonlinear, underactuated dynamics from incomplete, noisy sensor data. Recent advances in artificial intelligence (AI), particularly reinforcement learning (RL), have opened new perspectives for addressing such complex control problems through data-driven policy optimization from interaction with the environment. Yet purely data-driven methods are sample-inefficient, demanding extensive, sometimes unsafe exploration, especially without guiding physical models. This motivates hybrid AI–physics frameworks.
This article proposes a hybrid model-free/model-based flight-control approach using the reinforcement twinning algorithm. The model-based (MB) component uses an adjoint formulation and an adaptive digital twin continuously identified from live trajectories; the model-free (MF) component uses RL. The two agents share knowledge via transfer learning, imitation learning, and shared experience between the real environment and the digital twin, coordinated by a policy referee that selects which agent acts in reality based on digital-twin performance and a real-to-virtual consistency ratio.
The framework is evaluated for the longitudinal control of a flapping-wing drone, modelled as a nonlinear time-varying system driven by quasi-steady aerodynamic forces. The hybrid strategy is tested under three adaptive-model initializations: (1) offline identification from existing data, (2) random initialization with fully online identification, and (3) offline pre-training with biased parameters followed by online adaptation. In all cases, the hybrid framework improves performance, robustness, and sample efficiency over purely model-free and purely model-based approaches.
			
\vspace{7mm}
\end{abstract}

\section{Introduction}
\label{sec:1}
In recent years, the miniaturization of Unmanned Aerial Vehicle (UAV) technology has catalyzed the development of Micro Air Vehicles (MAVs). Unlike their larger counterparts, MAVs excel in confined and hazardous environments, from navigating collapsed structures for search and rescue to conducting industrial facility inspections and artificial pollination in dense vegetation \citep{Haider2021}. Future applications extend even beyond Earth’s frontiers, with MAVs currently being developed to spearhead the next generation of Martian exploration \citep{Pohly2021}.

However, at these small scales, viscous forces dominate inertial effects, severely degrading the aerodynamic efficiency and flight performance of fixed-wing and rotary-wing designs \citep{Mcmasters1980,Jones2010}.
Inspired by the superior flight performance of numerous insects and birds (\cite{Dudley2002,Chin2016}), flapping-wing micro air vehicles (FWMAVs) have emerged as a promising alternative. Pioneering work by \cite{Keennon2012} led to the first stable hovering flight of a hummingbird-like MAV, followed by a wide range of designs spanning centimeter- to decimeter-scale platforms \citep{Ma2013,Hsiao2024,Zhang2017,Karasek2018,Phan2017,Wang2024_ULB}. 
However, FWMAVs exhibit intrinsically unstable dynamics that require continuous and precise control of wing motion \citep{Menezes2016}. This remains a major challenge due to the FWMAV's strongly nonlinear, time-varying, and underactuated dynamics, further compounded by sensor noise, parametric uncertainties inherent to small-scale manufacturing (\cite{Wang2024_ULB}), and unsteady, poorly understood aerodynamic forces \citep{Poletti2024}.

Despite extensive research on FWMAV control \citep{Taha2012}, current approaches still fall short of replicating the exceptional capabilities of biological flyers \citep{Schenato2003,Taylor2002,Orlowski2012}. This work contributs to mitigating this gap proposing a hybrid control strategy in which a reinforcement learning policy and an adaptive digital twin interact to improve learning efficiency and robustness. To position this work, Section~\ref{sec1p1} reviews traditional model-based and learning-based control strategies for FWMAVs, while Section~\ref{sec1p2} discusses hybrid approaches that combine physics-based modeling and data-driven learning. The proposed framework and its specific contributions are then detailed in Section~\ref{sec1p3}.

\subsection{From Model-Based Control to AI-Driven Learning-Based Control}\label{sec1p1}

Most existing control strategies rely on traditional model-based (MB) approaches, typically using linear time-invariant (LTI) models or simplified analytical approximations of the FWMAV dynamics \citep{Tahmasian2014, Yao2019}. These models support well-established control techniques, including stability-based tools such as Routh-Hurwitz criteria and root locus methods \citep{Sun2007, Karasek2012, Aurecianus2020, Wang2024_ULB, Perez2011, Hu2024,Biswal2019}, as well as linear-quadratic regulator (LQR) \citep{Deng2006, Abbasi2022, Bhatia2014, Xiong2009}.

However, LTI models rely on simplified assumptions \citep{Taha2012}, which break down for FWMAVs with comparable wing-beat and body frequencies, complex kinematics, or agile manoeuvres \citep{Taha2012,Khosravi2021}. As a natural extension, control based on nonlinear formulations of the FWMAV dynamics has been explored using pseudo-inverse allocation \citep{Doman2010, Oppenheimer2011}, model-predictive control (MPC) \citep{Zhu2018, Zheng2024}, and central pattern generators \citep{Chung2010}. Reviews of these model-based approaches are provided in \cite{Taha2012,Orlowski2012}.

Despite these advances, most MB approaches remain inherently static and struggle to capture time-varying uncertainties arising from manufacturing tolerances, electromechanical wear, and measurement noise. Adaptive controllers have been proposed to mitigate these effects, for instance using Lyapunov-based adaptation \citep{Chirarattananon2014,Fei2023} or adaptive MPC formulations \citep{Zhu2017}. While these methods reduce model bias, they primarily focus on closed-loop response rather than continuously improving the underlying model. Similarly, \citet{Chand2016_2} employ recursive least squares (RLS) and adaptive pole placement control (APPC) to update an LTI model and its controller from flight data, but the approach remains limited by the linear model assumption.

Alternatively, model-free (MF) approaches avoid explicit modeling assumptions and rely on input–output data and trial-and-error exploration \citep{Vignon2023,Din2023,Pino2023,Razzaghi2023}. These methods naturally account for uncertainties captured in measurements and can interface effectively with nonlinear control laws \citep{Tedrake2009}.

Two broad categories of MF methods have been applied to FWMAVs. The first identifies system behaviour from data to adaptively tune a controller, including techniques such as model-free adaptive variable structure control (MFAVSC) \citep{Khosravi2021}, incremental proportional-integral-derivative control (iPID) \citep{Chand2016}, and fuzzy neural networks (FNN) \citep{Guo2008}. These approaches typically rely on local approximations, which limits their scalability across varying flight regimes. 

The second category adopts a purely trial-and-error paradigm, where reinforcement learning (RL) has emerged as the dominant artificial intelligence framework for data-driven control and biomimetic applications \citep{Verma2018, Zhu2022, Wang2024, Beckers2024, Novati2019, Lu2025}. Early applications of RL to flapping-wing systems focused on lift optimisation \citep{Motamed2007, Bayiz2019, Xiong2023}. More recently, \citet{Xue2023} used the soft actor-critic (SAC) algorithm to train a bumblebee model to hover in gusty conditions, while \citet{Fei2019, Tu2021} applied deep deterministic policy gradient (DDPG) to train a hummingbird-like robot and its digital twin to execute aggressive manoeuvres with strong similarity to biological flight. Nevertheless, the application of RL to flight control remains limited \citep{Giral2026}, primarily due to its poor sample efficiency and reliance on extensive trial-and-error exploration, with no guarantee of convergence \citep{Tu2021}.

\subsection{Hybrid Control Strategies: Integrating Physics-Based Models and Learning}\label{sec1p2}

Overall, model-based approaches are limited by model inaccuracies and uncertainties, while model-free methods suffer from poor sample efficiency and unstable training. This fundamental trade-off has motivated the development of hybrid strategies that combine physics-based modeling with data-driven learning.

In the context of FWMAVs, hybrid approaches remain scarce. \cite{Tu2021} is among the few contributions in this direction, using a reinforcement learning (RL) policy to assist or take over a model-based controller during highly dynamic manoeuvres.

More broadly, in the field of robotics and AI for dynamical systems, model-based reinforcement learning (MBRL) has gained increasing attention. These approaches leverage a variety of models, including physics-informed networks \citep{liu2021}, transformers \citep{Zhang2025}, and dynamical models derived from first principles \citep{lutter2021}. Such models are typically used to generate synthetic trajectories, accelerate policy optimization, warm-start RL training \citep{Qu2020}, or filter suboptimal policies \citep{Freed2024}. However, policy performance remains strongly dependent on model accuracy. In practice, many models fail to adapt online to evolving dynamics and are often optimized for global predictive accuracy rather than the specific regions of the state space that are most relevant for control \citep{Wang2019}.

To address model inaccuracies, system identification techniques have long been used in control to update model parameters from measurement data \citep{Ljung1999}. These approaches underpin adaptive control strategies such as self-tuning regulators, which adjust both model and controller parameters online \citep{astrom_adaptive_2008}. However, these methods are typically embedded within model-based control frameworks and do not fully exploit the potential of data-driven policy learning. In parallel, most existing MBRL approaches rely on a largely unidirectional interaction, where the model primarily supports the model-free (MF) policy optimization. \citet{Chebotar2017} proposed a more integrated approach by blending model-based and model-free policy updates, weighted by the discrepancy between real and predicted costs. Other hybrid strategies rely on policy switching mechanisms, where the best-performing controller is deployed while others continue to be trained. A notable example is the hierarchical mixtures of experts (HME) framework \citep{Yamada1997, Pinosky2023}, which favours model-based policies in regions of high MF uncertainty. Differently from what proposed in this work, these approaches do not seek to establish a bidirectional and synergistic interaction between model-based and model-free components. 

\subsection{Research Contributions and Article organization}\label{sec1p3}

Building upon recent advances in hybrid control and AI-driven learning for dynamical systems, this work introduces a reinforcement twinning framework tailored to the control of flapping-wing micro air vehicles (FWMAVs). Reinforcement twinning was recently proposed as a dual-agent model-based reinforcement learning framework enabling collaboration between model-free and model-based components through transfer learning, imitation learning, and shared experience \citep{Schena2023}. The present work extends this concept to the challenging context of FWMAV control, where strong nonlinearities, time-varying dynamics, and limited data availability require tightly coupled learning and modeling strategies. The proposed approach combines a model-free reinforcement learning policy with a model-based controller derived from an adaptive digital twin, enabling continuous interaction between data-driven learning and physics-based modeling.

A first contribution lies in the formulation of an online model identification strategy for FWMAV dynamics, where a simplified nonlinear time-invariant model is continuously calibrated from short-horizon trajectories generated by a nonlinear time-varying environment. This inverse modeling problem is addressed using a continuous adjoint-based optimization, allowing efficient parameter updates despite limited and evolving data. Unlike classical system identification approaches based on pre-collected datasets, the model is continuously updated from trajectories generated during control, allowing it to adapt to the regions of the state space effectively visited by the policy. A second contribution concerns the design of an improved bidirectional learning mechanism between the model-based and model-free components. In contrast to standard model-based reinforcement learning, where the model primarily supports policy optimization, the proposed framework enables reciprocal interactions through policy arbitration, experience sharing, and model-guided exploration. This coupling allows the model to guide the exploration of the control space, while the reinforcement learning agent enriches the data used for model identification.

Together, these elements define a hybrid learning framework that improves sample efficiency, robustness, and adaptability in the control of FWMAVs, particularly in the presence of model uncertainty and evolving dynamics.

\section{Problem definition}\label{sec_def}

\subsection{Canonical FWMAV configuration}

The control problem considered in this work is to determine the flapping kinematics that generate the aerodynamic forces required to steer a flapping-wing micro air vehicle (FWMAV) toward a prescribed target state within a short time horizon. The objective is to define a benchmark that captures the main nonlinear and time-varying features of FWMAV flight while remaining computationally tractable for repeated policy-learning episodes.

\begin{figure*}
    \centering
    \begin{subfigure}{0.43\textwidth}
        \centering
        \includegraphics[width=\textwidth]{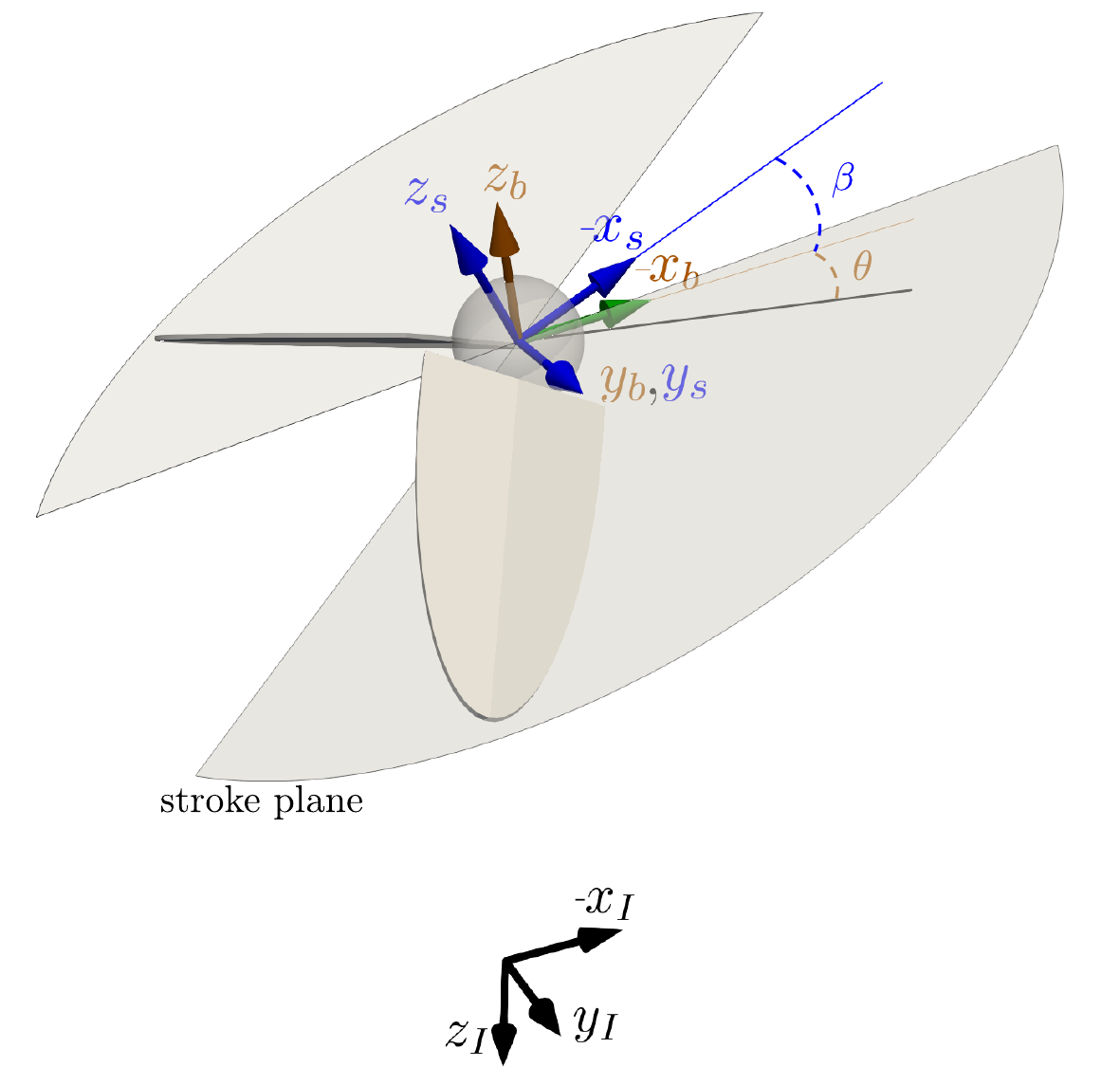}
        \caption{}
        \label{fig_drone1}
    \end{subfigure}
    \hfill
    \begin{subfigure}{0.53\textwidth}
        \centering
        \includegraphics[width=0.99\textwidth]{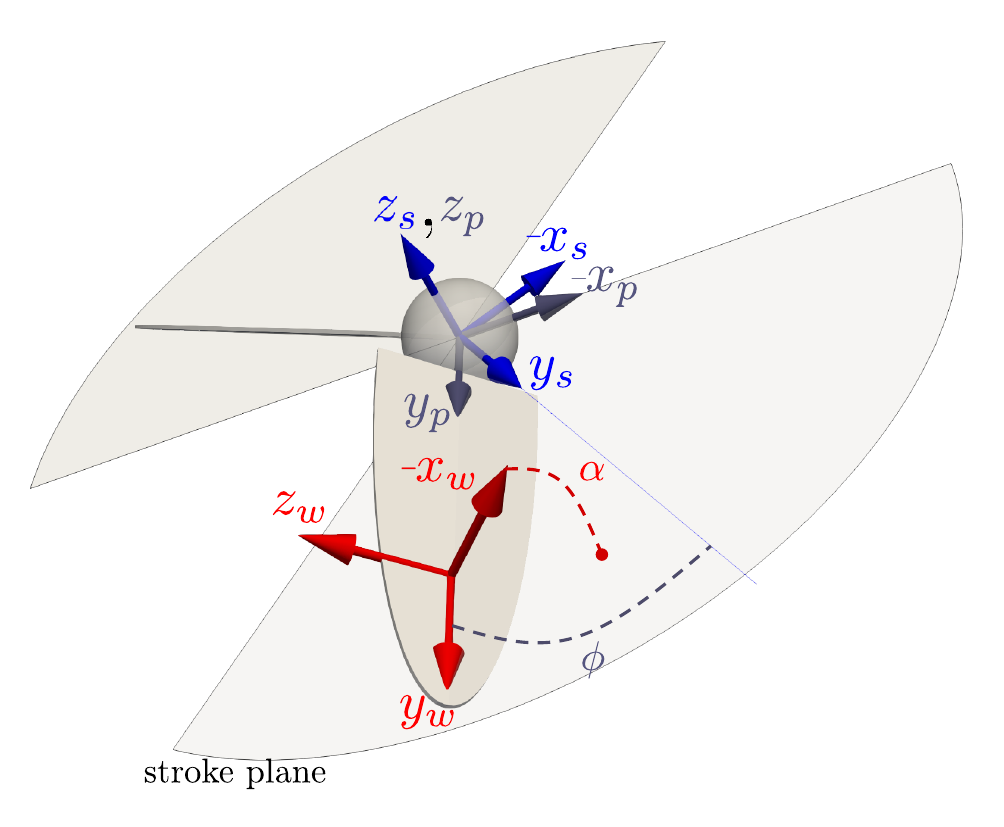}
        \caption{}
        \label{fig_drone2}
    \end{subfigure}
    \caption{(a) Schematic of the flapping-wing drone control problem and (b) focus on the wing kinematics defined by the flapping angle $\phi$ and the pitching angle $\alpha$}
    \label{fig_drone12}
\end{figure*}

To this end, we consider a canonical flapping-wing drone that facilitates the development and assessment of the reinforcement twinning framework while preserving the key physical characteristics of realistic FWMAV prototypes. The drone is modeled as a spherical body equipped with two semi-elliptical, rigid, and massless wings, consistently with previous works \citep{Calado2023,Schena2023,Poletti2024} (Figure~\ref{fig_drone1}). The wings have mean chord length $c=1$ cm, span $R=5$ cm, and thickness $b=0.03c$. The rigid body has mass $m_b=3$ g, radius $R_b=c$, and moment of inertia $I_b=1.2\times10^{-5}\,\mathrm{kg\,m^2}$. The body center is taken as the instantaneous center of rotation, while the wing root is located 2.25 cm away along the spanwise direction.

\subsection{Control objective and episodic formulation}

We focus on vertical ascending flights under the assumption of bilateral symmetry of the flapping wings. Following \cite{Yao2018}, the problem is reduced to the longitudinal dynamics of the drone, characterized by three degrees of freedom in the inertial reference frame $(xyz)_I$: horizontal translation $x$, vertical translation $z$, and pitch angle $\theta$, defined between the body-fixed axis $x_b$ and $x_I$. Yaw, roll, and lateral displacement are neglected.

The system state is defined as
\[
\bm{s} = [\dot{x},\dot{z},\dot{\theta},x,z,\theta]^T.
\]
Starting from rest at the origin ($\bm{s}_0=\bm{0}$), the objective is to reach the target state
\[
\tilde{\bm{s}} = [0,0,0,0,1,0]^T\,,
\]
within a finite time horizon $T_0 = 1.5$ s. This duration is consistent with fast vertical maneuvers observed in natural flyers such as hummingbirds \citep{Ortega2018}, and provides a challenging yet computationally affordable benchmark.

Following a traditional reinforcement learning setting, the control policy is evaluated over repeated trials, referred to as \emph{episodes}, each defined over the time interval $t \in [0,T_0]$. During an episode, the system trajectory is sampled once per flapping cycle at frequency $f_c$, resulting in $N_c = f_c T_0$ discrete time steps. The time instants are defined as $t_k = k\Delta t$ with $\Delta t = 1/f_c$. At each cycle, a cycle-averaged state is computed as
\begin{equation}
\bm{s}_k = f_c \int^{k/f_c}_{(k-1)/f_c} \bm{s}(t)\,dt,
\end{equation}
and the tracking error is defined as
\begin{equation}
\bm{e}_k = \bm{s}_k - \tilde{\bm{s}}.
\label{err}
\end{equation}

\subsection{Reward definition}

The control performance is quantified through a reward function defined at each cycle. The tracking error is decomposed into velocity and position components:
\[
\bm{e}_{vel,k} \in \mathbb{R}^3, \quad \bm{e}_{pos,k} \in \mathbb{R}^3,
\]
corresponding to the first and last three entries of $\bm{e}_k$, respectively. The reward is defined as
\begin{equation}
r_k(\bm{e}_k) = -\|\bm{e}_{pos,k}\|_2^2 - \eta \|\bm{e}_{vel,k} \circ \bm{h}(\bm{e}_{pos,k})\|_2^2,
\label{eq_rk}
\end{equation}
where $\circ$ denotes the Hadamard (element-wise) product and $\eta$ is a weighting coefficient.

The function $\bm{h}(\bm{e}_{pos,k})$ modulates the contribution of the velocity error, reducing its influence when the system is far from the target and increasing it as the target is approached. This behavior is implemented using a vector of Gaussian functions:
\begin{equation}
\bm{h}(\bm{e}_{pos,k}) = \exp\left(-\bm{e}_{pos,k} \circ \bm{e}_{pos,k} \circ \frac{1}{\bm{\sigma}^2}\right),
\end{equation}
where $\bm{\sigma} = [\sigma_x,\sigma_z,\sigma_\theta]$ controls the spatial decay of the weighting. The exponential is applied element-wise.

\subsection{Control parametrization}

To maximize the reward, the drone adapts its flapping kinematics through a low-dimensional control parametrization. Actuation dynamics and delays are neglected in this study to isolate the control-learning problem, which is a common assumption in preliminary investigations of flapping-wing control \citep{Schenato2003,Yao2019,Faruque2018}.

The wing motion is described using three Euler angles (Figure~\ref{fig_drone12}):
(i) the stroke plane angle $\beta$, defined between $x_b$ and $x_s$;
(ii) the flapping angle $\phi$, defined between $y_p$ and $y_s$;
and (iii) the pitching angle $\alpha$, defined between $x_p$ and the chord-normal direction $x_w$ of the wing-attached frame.

The flapping and pitching motions follow harmonic laws:
\begin{align}
\phi(t) &= A_\phi \cos(2\pi f_c t) + A_{\text{off}}, \label{eq_phi}\\
\alpha(t) &= A_\alpha \sin(2\pi f_c t), \label{eq_alpha}
\end{align}
where $f_c = 20$ Hz is the flapping frequency and $A_\alpha = 45^\circ$ is the pitching amplitude.

The control input is defined as
\[
\bm{a} = [A_\phi, \beta, A_{\text{off}}]^T,
\]
which governs the wing kinematics at the cycle level.

The control policy is parameterized as a clipped and scaled proportional-derivative (PD) mapping:
\begin{equation}
\pi(\bm{e}_k;\bm{w}_a) =
\frac{\tanh(\bm{W}_a \bm{e}_k + \bm{b}) + 1}{2}
(\bm{a}_{max} - \bm{a}_{min}) + \bm{a}_{min},
\label{eq_pi}
\end{equation}
where $\bm{W}_a \in \mathbb{R}^{3\times6}$ is the weight matrix, $\bm{b} \in \mathbb{R}^3$ is a bias vector, and $\bm{w}_a$ denotes the vectorized form of $\bm{W}_a$.

The action bounds are defined as
\[
\bm{a}_{max} = [88^\circ,\,30^\circ,\,0.5^\circ], \quad
\bm{a}_{min} = [50^\circ,\,-30^\circ,\,-0.5^\circ],
\]
based on observations of natural flyers \citep{Ellington1984_III}. Compared to standard formulations \citep{Yao2019,Faruque2018,Karasek2012}, this parametrization includes cross-coupling and bias terms, which improve robustness to modeling errors and disturbances \citep{Chirarattananon2014,Tu2021_2}.

Finally, ideal state estimation is assumed, neglecting sensor dynamics and noise. This simplification allows the study to focus on the interaction between control learning and system dynamics, and is consistent with prior work in this area.

\section{Reinforcement twinning for hybrid control learning}\label{sec_meth}

The reinforcement twinning (RT) algorithm is used to train the control policy defined in equation~\eqref{eq_pi}. The method was first introduced in \cite{Schena2023}, to which the reader is referred for a more complete presentation. Here, we provide a concise description tailored to the present FWMAV control problem, with emphasis on the general principle (Section~\ref{eq_generalPrinciple}), the model-free and model-based optimization loops (Sections~\ref{sec_ddpg} and \ref{sec_adjoint}), and the proposed cooperation mechanisms that couple the two learning processes (Section~\ref{sec_coop}).    

\subsection{General principle}\label{eq_generalPrinciple} 

\begin{figure*}
    \centering
    \includegraphics[width=0.95\linewidth]{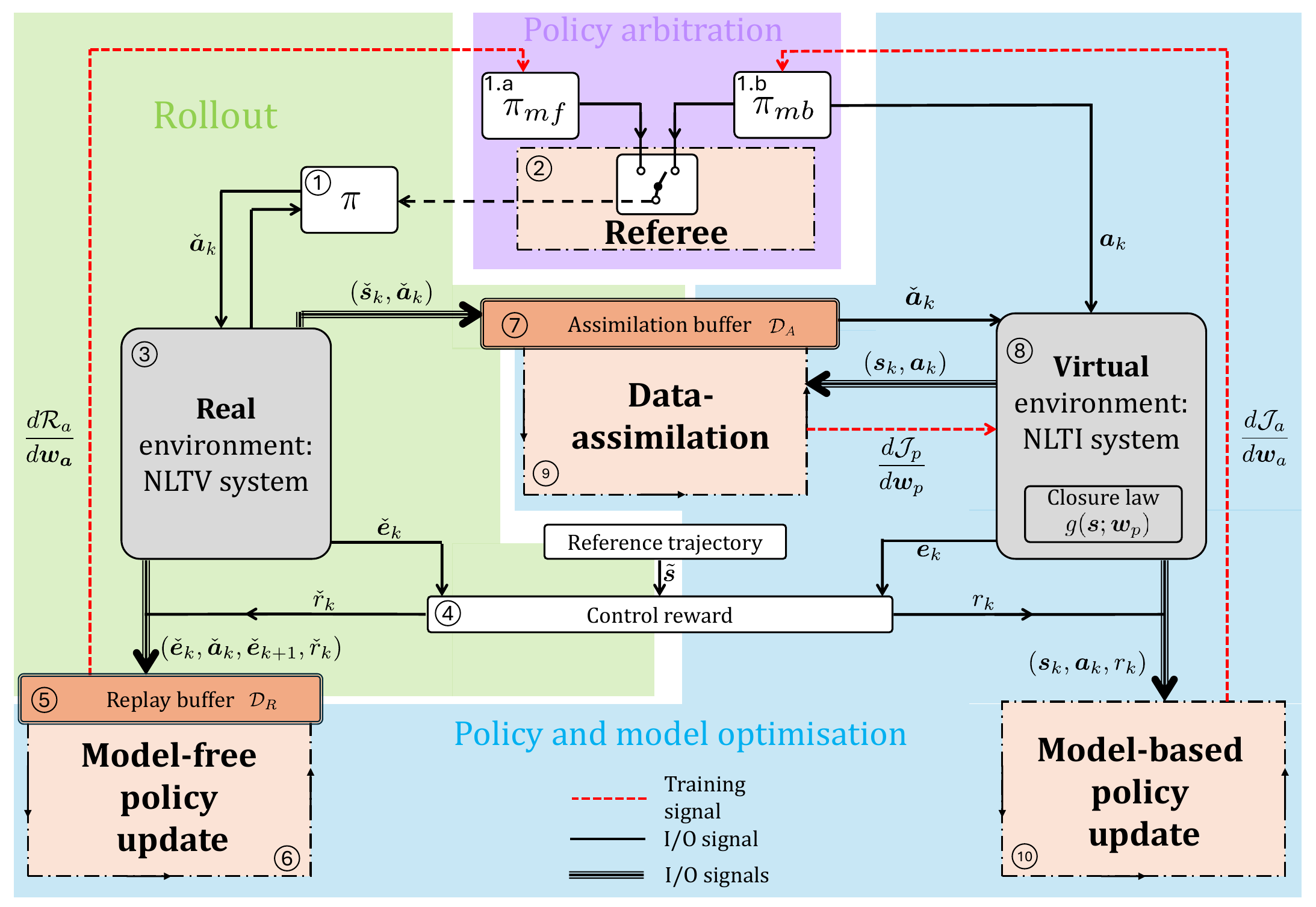}
    \caption{Block diagram of the reinforcement twinning algorithm hybridizing a model-free (1.a) and model-based (1.b) policy to train a control agent $\pi$ (1.) using a virtual environment (8.) assimilated from live data of the real environment (3.).}
    \label{fig_diag}
\end{figure*}

\begin{algorithm}
\caption{Reinforcement twinning (RT) Algorithm }\label{alg:main}
Initialize the algorithm's weights including the critic ($\bm{w}_{q}$), the model-free policy ($\bm{w}_{a,mf}$), the model-based policy ($\bm{w}_{a,mb}$), and the closure law ($\bm{w}_p$). 
\begin{algorithmic}
\Procedure{RT Updates}{$\bm{w}_q$, $\bm{w}_{a,mf}$, $\bm{w}_p$, $\bm{w}_{mb}$}
    \State \parbox[t]{\dimexpr\linewidth-\algorithmicindent}{\textbf{Step 1: Collect} $N_c$ transitions into $\mathcal{D}_{R}$ and $N_c$ state-action pairs into $\mathcal{D}_{A}$ as the live policy $\pi(\check{\bm{e}};\bm{w}_a)$ interacts with the real environment for one episode.\par}
    
    \State \parbox[t]{\dimexpr\linewidth-\algorithmicindent}{\textbf{Step 2: Update} the critic weights $\bm{w}_q$ and policy weights $\bm{w}_{a,mf}$ using the DDPG algorithm (6) and mini-batches from $\mathcal{D}_{R}$.\par}
    
    \State \parbox[t]{\dimexpr\linewidth-\algorithmicindent}{\textbf{Step 3: Replay} the $n_e$ episodes from $\mathcal{D}_{A}$ on the virtual environment, collect the associated virtual state-action pairs and compute the assimilation cost.\par}
    
    \State \parbox[t]{\dimexpr\linewidth-\algorithmicindent}{\textbf{Step 4: Update} the virtual environment closure law $\bm{w}_p$ using the adjoint-based data-assimilation loop (9).\par}
    
    \State \parbox[t]{\dimexpr\linewidth-\algorithmicindent}{\textbf{Step 5: Collect} $N_c$ state-action pairs and compute the associated control cost as the model-based policy $\pi(\bm{e};\bm{w}_{a,mb})$ interacts with the virtual environment.\par}
    
    \State \parbox[t]{\dimexpr\linewidth-\algorithmicindent}{\textbf{Step 6: Update} the policy weights $\bm{w}_{a,mb}$ using the adjoint-based data-assimilation loop (10) and teh associated virtual state-action pairs.\par}
    
    \State \parbox[t]{\dimexpr\linewidth-\algorithmicindent}{\textbf{Step 7: Evaluate} the best-performing policy, deploy it live, and enable policy collaboration through the policy referee (2).\par}
\EndProcedure
\end{algorithmic}
\end{algorithm}

Figure~\ref{fig_diag} summarizes the main components of the RT framework, labelled from (1) to (10). Two control policies, both parameterized according to equation~\eqref{eq_pi}, are trained in parallel: a model-free policy $\pi_{mf}(\bm{e}_k;\bm{w}_{a,mf})$ (1.a) and a model-based policy $\pi_{mb}(\bm{e}_k;\bm{w}_{a,mb})$ (1.b). Their interaction is governed by a referee (2), described in Section~\ref{sec_coop}, which selects which policy is deployed in the real environment (3). The selected policy is referred to as the \emph{live policy}, denoted $\pi(\bm{e}_k;\bm{w}_a)$.

The real environment emulates the FWMAV dynamics through the nonlinear time-varying model introduced in Section~\ref{sec_NLTP}. This environment is assumed unknown to the RT algorithm. Only the cycle-wise rewards $\check{r}_k$ (equation~\eqref{eq_rk}), cycle-averaged tracking errors $\check{\bm{e}}_k$ (equation~\eqref{err}), and applied control actions $\check{\bm{a}}_k$ (equation~\eqref{eq_pi}) are available to the learning framework. These quantities are stored in two memory buffers after each flapping cycle.

The replay buffer $\mathcal{D}_{R}$ (5), with capacity $n_{R}$, stores transition tuples of the form
\((\check{\bm{e}}_k,\check{\bm{a}}_k,\check{\bm{e}}_{k+1},\check{r}_k),\)
while the assimilation buffer $\mathcal{D}_{A}$ (7), with capacity $n_A$, stores cycle-averaged state--action trajectories. Throughout the paper, the check notation indicates quantities collected from the real environment.

The two buffers play distinct roles and therefore follow different update strategies. The replay buffer $\mathcal{D}_{R}$ is managed using a first-in, first-out (FIFO) rule. By contrast, the assimilation buffer $\mathcal{D}_{A}$ retains the $n_e=n_A/N_c$ trajectories that maximize a variance criterion, thereby promoting diversity in the state--action pairs used for model learning. This variance is computed from the trajectory matrices $\bm{S}_i \in \mathbb{R}^{N_c\times 6}$, defined as
\[
\bm{S}_i = \operatorname{concat}(\check{\bm{s}}_0,\ldots,\check{\bm{s}}_{N_c}),
\]
where $\operatorname{concat}$ stacks the state vectors over one episode. The buffer variance is then written as
\begin{equation}
    \xi = \frac{1}{n_A}\sum_{i=1}^{n_e}\operatorname{Tr}\!\left[(\overline{\bm{S}}-\bm{S}_i)(\overline{\bm{S}}-\bm{S}_i)^T\right],
\end{equation}
where
\[
\overline{\bm{S}}=\frac{1}{n_e}\sum_{i=1}^{n_e}\bm{S}_i
\]
is the mean trajectory matrix and $\operatorname{Tr}$ denotes the trace operator. The role of this diversity criterion is assessed in Section~\ref{sec_results}, where it is compared against a minimum-error strategy that instead retains the trajectories closest to the target.

After the rollout stage (green-shaded region in Figure~\ref{fig_diag}), the replay and assimilation buffers are used in two coupled optimization loops (blue-shaded region). In the first loop (6), mini-batches sampled from the replay buffer $\mathcal{D}_{R}$ are used to update the model-free policy $\pi_{mf}(\bm{e}_k;\bm{w}_{a,mf})$ through the deep deterministic policy gradient (DDPG) algorithm, described in Section~\ref{sec_ddpg}.

In parallel, the $n_e$ trajectories stored in the assimilation buffer $\mathcal{D}_{A}$ are used to identify a simplified model of the real environment. This virtual environment (8), described in Section~\ref{sec_NLTI}, is a computationally efficient nonlinear time-invariant model defined through a closure law $\bm{p}=g(\bm{s},\bm{a};\bm{w}_{p})$. The closure is identified through an adjoint-based data-assimilation procedure (9), described in Section~\ref{sec_adjoint}. Once updated, the virtual environment is used to optimize the model-based policy $\pi_{mb}(\bm{e}_k;\bm{w}_{a,mb})$ through a second adjoint-based loop (10).

After both policy updates, the referee regulates the interaction between the two policies and selects which one will be deployed as the live policy during the next roll-out. This policy-arbitration stage is shown in purple in Figure~\ref{fig_diag}. In summary, reinforcement twinning combines real-environment interaction, adaptive model learning, model-free reinforcement learning, and model-based optimization within a single hybrid learning architecture. Algorithm~\ref{alg:main} summarizes the overall workflow.

\subsection{Model-free policy updates}\label{sec_ddpg}

After each interaction of the live policy with the real environment (Step 1 of Algorithm~\ref{alg:main}), the model-free optimization loop ((6) in Figure~\ref{fig_diag}) updates the policy so as to maximize the cumulative reward collected over the episode:
\begin{equation}
    \check{\mathcal{R}}_a = \sum_{k=0}^{N_c-1}\check{r}_k(\check{\bm{e}}_k,\check{\bm{a}}_k),
\label{eq_Ra}
\end{equation}
where $\check{r}_k$ is the instantaneous reward defined in equation~\eqref{eq_rk}.

This optimization relies on the deep deterministic policy gradient (DDPG) algorithm \citep{Lillicrap2015}, an off-policy actor--critic method for continuous control. In DDPG, the policy update is guided by a critic neural network $q(\bm{e}_k,\bm{a}_k;\bm{w}_q)$, which approximates the action-value function, or Q-value, namely the expected cumulative reward associated with a state--action pair under the current policy \citep{Sutton2018}.

In the present work, the critic network contains two hidden layers of 256 neurons with ReLU activation functions and is parameterized by the weights $\bm{w}_q$. The critic is trained by minimizing the mean squared Bellman error using mini-batches sampled from the replay buffer $\mathcal{D}_{R}$, together with a target critic network with weights $\bm{w}_{q'}$ \citep{Lillicrap2015}. The target network has the same architecture as the critic and is updated through the soft-update rule
\(
\bm{w}_{q'}=\zeta \bm{w}_{q}+(1-\zeta)\bm{w}_{q'},
\)
where $\zeta$ is a small relaxation parameter that improves training stability.

The critic gradient is then used to compute the policy gradient driving the update of the model-free actor:
\begin{equation}
    \frac{\partial \check{\mathcal{R}}_a}{\partial \bm{w}_{a,mf}}
    =
    \mathbb{E}_{\sim \mathcal{D}_{R}}
    \left\{
    \frac{\partial q(\check{\bm{e}},\check{\bm{a}};\bm{w}_q)}{\partial \check{\bm{a}}}
    \frac{\partial \pi_{mf}(\check{\bm{e}};\bm{w}_{a,mf})}{\partial \bm{w}_{a,mf}}
    \right\},
\label{eq_policy_grad}
\end{equation}
where $\mathbb{E}$ denotes the empirical average over the sampled mini-batch.

Compared with the original DDPG formulation, the critic and actor are updated $n_q$ and $n_a$ times after each real-environment episode, respectively, in order to maintain a similar optimization cadence to that used in the model-based branch. In Algorithm~\ref{alg:main}, these operations correspond to Step 2.

\subsection{Model-based policy updates}\label{sec_adjoint}

After the model-free update, the model-based branch performs two successive optimization (learning) tasks:  
(i) update the closure law $g(\bm{s},\bm{a};\bm{w}_{p})$ of the virtual environment, and  
(ii) optimizing the model-based policy $\pi_{mb}(\bm{e};\bm{w}_{a,mb})$ within that environment.  
These operations correspond to Steps 3 to 6 in Algorithm~\ref{alg:main}.

The first task uses the $n_e$ trajectories stored in the assimilation buffer $\mathcal{D}_{A}$. The control actions recorded in these trajectories are replayed in the virtual environment, and the resulting virtual states $\bm{s}$ are compared with the real states $\check{\bm{s}}$ stored in the buffer. This yields the assimilation cost
\begin{equation}
\begin{split}
    \mathcal{J}_p(\bm{s},\check{\bm{s}},\bm{w}_p) = {} & \mathbb{E}_{\sim \mathcal{D}_{A}} \Biggl\{ \int_0^{T_0} \Bigl[ \|\bm{s}(t)-\check{\bm{s}}(t)\|_2^2 \\
    & + \alpha_p \|\bm{w}_p\|_2^2 \Bigr] dt \Biggr\},
\end{split}
\label{eq_Jp}
\end{equation}
where the first term measures the mismatch between real and virtual trajectories and the second term regularizes the closure-law parameters, promoting robustness across diverse flight conditions \citep{Calado2023}.

The closure parameters are updated using gradients computed through the continuous adjoint method. For each replayed episode, the corresponding adjoint variable $\bm{\lambda}_p$ satisfies the terminal-value problem
\begin{equation}
\begin{cases}
 \dot{\bm{\lambda}}_p = -\left(\frac{\partial f}{\partial \bm{s}}\right)^T\bm{\lambda}_p - \left(\frac{\partial \mathcal{J}_p}{\partial \bm{s}}\right)^T,\\[4pt]
 \bm{\lambda}_p(T_0)=\bm{0},
\end{cases}
\label{lambda_p}
\end{equation}
where $f(\bm{s},\bm{a},\bm{w}_p)$ denotes the flow map of the virtual environment introduced in Section~\ref{sec_NLTI}. The derivatives $\partial f/\partial \bm{s}$ and $\partial \mathcal{J}_p/\partial \bm{s}$ are obtained symbolically from the analytical model.

The resulting gradient of the assimilation cost with respect to the closure parameters is
\begin{equation}
\frac{d \mathcal{J}_p}{d \bm{w}_p}
=
\mathbb{E}_{\sim \mathcal{D}_{A}}
\left\{
\int_0^{T_0}
\left[
\bm{\lambda}_p^T\frac{\partial f}{\partial \bm{w}_p}
+
\frac{\partial \mathcal{J}_p}{\partial \bm{w}_p}
\right]dt
\right\}.
\label{eq_dpdJ}
\end{equation}

This gradient is used in an ADAM optimization loop \citep{Kingma2014}, run for $n_g$ iterations, to update $\bm{w}_p$ (Step 4 of Algorithm~\ref{alg:main}).

Once the virtual environment has been updated, it is used to optimize the model-based policy. For a rollout generated by the current model-based controller, the associated control objective is written in continuous form as
\begin{equation}
    \mathcal{J}_a(\bm{s},\tilde{\bm{s}},\bm{w}_{a,mb})
    =
    \int_0^{T_0} r(t)\,dt,
\label{eq_Ja}
\end{equation}
where $r(t)$ is the instantaneous reward associated with the virtual trajectory.

The policy gradient is again obtained through an adjoint formulation:
\begin{equation}
\frac{\partial \mathcal{J}_a}{\partial \bm{w}_{a,mb}}
=
\int_0^{T_0}
\bm{\lambda}_a^T
\frac{\partial f}{\partial \bm{a}}
\frac{\partial \pi_{mb}}{\partial \bm{w}_{a,mb}}
\,dt,
\label{eq_dpdR}
\end{equation}
with
\begin{equation}
\begin{cases}
 \dot{\bm{\lambda}}_a = -\left(\frac{\partial f}{\partial \bm{s}}\right)^T\bm{\lambda}_a - \left(\frac{\partial \mathcal{J}_a}{\partial \bm{s}}\right)^T,\\[4pt]
 \bm{\lambda}_a(T_0)=\bm{0}.
\end{cases}
\end{equation}

This gradient is then used in a second ADAM loop, run for $n_{mb}$ iterations, to update the weights $\bm{w}_{a,mb}$ (Step 6 of Algorithm~\ref{alg:main}).

\subsection{The learning feedback between model-free and model-based}\label{sec_coop}

\begin{figure*}
    \centering
    \includegraphics[width=0.74\linewidth]{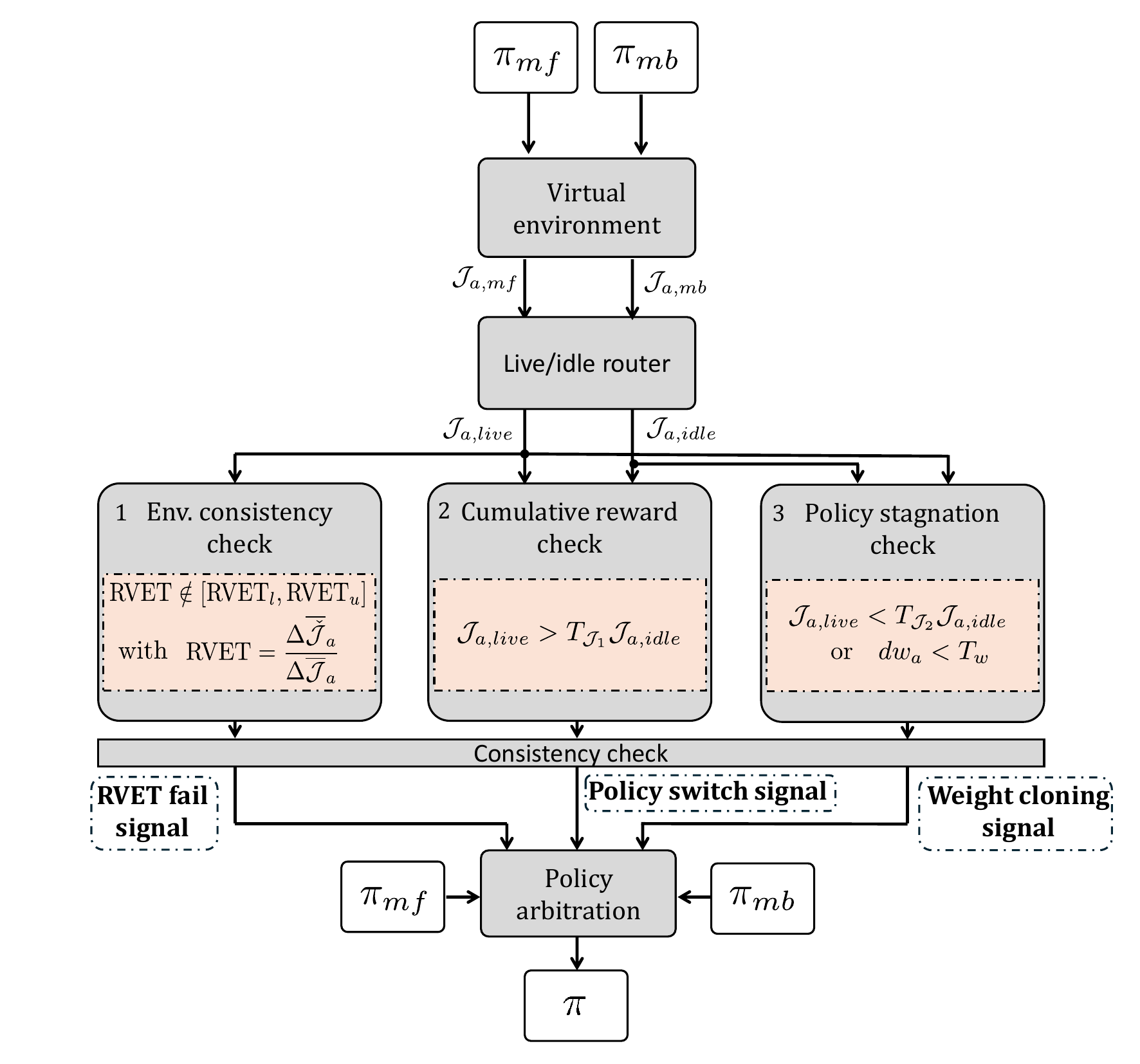} %
    \caption{Block diagram of the policy referee showing the cooperation mechanisms between the model-based and model-free policy used to define the live policy}
    \label{fig_diag2}
\end{figure*}

After both optimization of $\pi_{mf}$ (Step 2) and $\pi_{mb}$ (Step 6) have been updated, the interaction between the two policies is regulated by a policy referee, shown in Figure~\ref{fig_diag2}. Each policy is first evaluated for one episode in the virtual environment, which provides the corresponding control costs according to equation~\eqref{eq_Ja}. These costs are labelled $\mathcal{J}_{a,live}$ and $\mathcal{J}_{a,idle}$ depending on whether the policy under consideration was or was not deployed in the real environment during the preceding roll-out.

These live and idle costs are then processed by three cooperation mechanisms, labelled (1) to (3) in Figure~\ref{fig_diag2}.

The first mechanism assesses the reliability of the virtual environment for policy evaluation. This is done through the real-to-virtual environment trust ratio (RVET), inspired by surrogate-based optimization \citep{Koziel2011} and trust-region methods \citep{Nocedal1999}. For two successive episodes $i-1$ and $i$, the RVET is defined as
\begin{equation}
    \mathrm{RVET}_i
    =
    \frac{
    \overline{\check{\mathcal{J}}}_{a,i}-\overline{\check{\mathcal{J}}}_{a,i-1}
    }{
    \overline{\mathcal{J}}_{a,i}-\overline{\mathcal{J}}_{a,i-1}
    }
    =
    \frac{\Delta \overline{\check{\mathcal{J}}}_{a}}{\Delta \overline{\mathcal{J}}_{a}},
\label{eq_MET}
\end{equation}
where the overbar denotes a moving average over recent episodes and the \emph{live} subscript is omitted for brevity. The quantity measures whether improvements predicted in the virtual environment are consistent with those observed in the real environment. The RVET must remain above a lower threshold $\mathrm{RVET}_l$ to preserve the sign of the predicted improvement, and below an upper threshold $\mathrm{RVET}_u$ to ensure sufficient similarity between real and virtual responses.

If either condition is repeatedly violated, the referee raises an RVET-fail signal and enforces the model-free policy as the live policy. In this regime, the model-based policy is temporarily prevented from controlling the real environment, and its training is paused until the virtual environment regains sufficient reliability.

The second mechanism (central branch in Figure \ref{fig_diag2}) compares the live and idle policies through their virtual-environment costs. If the live policy consistently underperforms the idle one according to the threshold $T_{\mathcal{J}_1}$, the referee issues a policy-switch signal and the previously idle policy becomes live. This mechanism allows the framework to deploy the better-performing controller while avoiding premature switching through an additional consistency check.

The third mechanism (right branch in Figure \ref{fig_diag2}) detects stagnation of the idle policy. If the normalized weight update $dw_a$ falls below the threshold $T_w$, suggesting that learning is trapped in a poor local minimum, or if the live policy consistently and significantly outperforms the idle policy according to the threshold $T_{\mathcal{J}_2}$, the referee triggers a weight-cloning signal. Depending on which policy is live, the weights are then copied from the better-performing policy to the weaker one. This mechanism prevents excessive divergence between the two branches and accelerates the exploration of promising regions of the policy space.

Taken together, these three mechanisms define the bidirectional cooperation between model-free and model-based learning. In addition, interaction through the real environment induces an indirect exchange of information between the two branches. When the model-based policy is deployed and performs well, its transitions are inserted into the replay buffer $\mathcal{D}_{R}$ and contribute to the training of the critic and of the model-free policy. Conversely, when the model-free policy is deployed, the resulting trajectories can enrich the assimilation buffer and improve the identification of the virtual environment. In this way, the two learning processes do not simply coexist; they continuously support each other through shared experience, adaptive arbitration, and model-guided exploration.

\section{Real and Virtual Environments}\label{sec_model}

\subsection{Real environment: nonlinear time-varying (NLTV) dynamics}\label{sec_NLTP}

The real environment simulates the flight dynamics of the flapping-wing drone introduced in Section~\ref{sec_def}. To keep the benchmark computationally affordable while preserving the dominant body dynamics, the multibody equations of motion are simplified by neglecting the wing inertia \citep{Samin2003,Sun2007_2}. Under this assumption, the system reduces to the dynamics of a rigid body subjected to aerodynamic loads and gravity applied at the barycenter \citep{Etkin1995}. This approximation is commonly adopted in control-oriented studies of flapping flight \citep{Taha2012}.

Restricting attention to the longitudinal dynamics, the Newton--Euler equations in the body-fixed frame read
\begin{equation}
\label{eq_instSystem}
    \begin{bmatrix}
        \ddot{x}^b \\
        \ddot{z}^b \\
        \ddot{\theta}^b
    \end{bmatrix}
    =
    \begin{bmatrix}
        F^b_x/m_b \\
        F^b_z/m_b \\
        T^b_y/I_b
    \end{bmatrix}
    +
    \begin{bmatrix}
        -g\sin(\theta^b)-\dot{\theta}^b\dot{z}^b \\
        g\cos(\theta^b)+\dot{\theta}^b\dot{x}^b \\
        0
    \end{bmatrix},
\end{equation}
where $(F_x^b,F_z^b,T_y^b)$ are the aerodynamic forces and pitch moment expressed in the body frame. Since these loads are evaluated instantaneously within each wingbeat cycle, equation~\eqref{eq_instSystem} defines a nonlinear time-varying (NLTV) system.

Introducing the state vector
\[
\bm{s}^b = [\dot{x}^b,\dot{z}^b,\dot{\theta}^b,x^b,z^b,\theta^b]^T,
\]
equation~\eqref{eq_instSystem} can be written compactly as
\begin{equation}
\label{eq_fs}
    \frac{d \bm{s}^b}{d t}
    =
    \bm{g}_1(\bm{s}^b,\bm{a},t)+\bm{g}_2(\bm{s}^b),
\end{equation}
where $\bm{g}_1$ gathers the time-varying aerodynamic contributions and $\bm{g}_2$ contains the remaining inertial and gravitational terms.

The aerodynamic loads are estimated using the quasi-steady model of \cite{Lee2016}, in which the total force is decomposed into four contributions. In the present work, only the dominant component is retained, as it largely outweighs the others for the smooth wing kinematics considered here (equations~\eqref{eq_phi}--\eqref{eq_alpha}) \citep{Lee2016}. This contribution captures the effect of the leading-edge vortex (LEV), which remains attached to the leeward side of the wing at high angles of attack \citep{Eldredge2019}.

Using a blade-element formulation, the instantaneous lift $L(t)$ and drag $D(t)$ are written as
\begin{equation}
\label{eq_lift}
\begin{bmatrix}
L(t) \\
D(t)
\end{bmatrix}
=
\frac{1}{2}\rho
\begin{bmatrix}
C_L(\alpha_e) \\
C_D(\alpha_e)
\end{bmatrix}
\int_{\Delta R}^{\Delta R+R}
\|\bm{U}_w(t,r)\|_2^2\,c(r)\,dr,
\end{equation}
where $\rho$ is the air density. Here, the drag is aligned with the relative wing velocity $\bm{U}_w$, while the lift is normal to it.

The relative velocity $\bm{U}_w$ combines the contribution of wing flapping, body pitching, and body translation. Expressed in the wing-attached frame $(xyz)^w$, it reads
\begin{equation}
\label{drone_eq_Uw}
\begin{aligned}
\bm{U}_w(r,t,\bm{u})
&=
\bm{R}^w_y(\alpha)
\Bigg[
\Bigg(
\bm{R}^p_z(\phi)\bm{R}^b_y(\beta)
\begin{bmatrix}
0 \\
\dot{\theta}^b \\
0
\end{bmatrix}
+
\begin{bmatrix}
0 \\
0 \\
\dot{\phi}
\end{bmatrix}
\Bigg)
\times
\begin{bmatrix}
0 \\
r \\
0
\end{bmatrix}
\Bigg]
\\
&\quad
+
\bm{R}^w_y(\alpha)\bm{R}^p_z(\phi)\bm{R}^b_y(\beta)
\begin{bmatrix}
\dot{x}^b \\
0 \\
\dot{z}^b
\end{bmatrix}.
\end{aligned}
\end{equation}
Here, $\bm{R}$ denotes a rotation matrix, with the subscript indicating the rotation axis and the superscript identifying the reference frame. The full set of rotation matrices is reported in \cite{Schena2023,Cai2021}.

The aerodynamic coefficients in equation~\eqref{eq_lift} depend on the effective angle of attack
\[
\alpha_e = \cos^{-1}\!\left(\frac{U_{w,z}}{\|\bm{U}_w\|_2}\right),
\]
defined as the angle between the chord-normal direction $z^w$ (Figure~\ref{fig_drone2}) and the relative wing velocity $\bm{U}_w$. Following \cite{Lee2016}, the lift and drag coefficients are modeled as
\begin{align}
    C_L &= a\sin(2\alpha_e), \label{drone_eq_Clift}\\
    C_D &= b + c\left(1-\cos(2\alpha_e)\right), \label{drone_eq_Cdrag}
\end{align}
where $a$, $b$, and $c$ are obtained from empirical correlations fitted to CFD simulations over a broad range of Reynolds numbers, Rossby numbers, wing aspect ratios, and taper ratios \citep{Lee2016}.

The aerodynamic loads are finally projected onto the body frame as
\begin{equation}
\label{eq_F}
\begin{bmatrix}
F^b_x \\
0 \\
F^b_z
\end{bmatrix}
=
\bm{R}^b_y(-\beta)\bm{R}^p_z(-\phi)\bm{R}^w_y(-\alpha)\bm{R}_y(\alpha_e)
\begin{bmatrix}
D \\
0 \\
L
\end{bmatrix},
\end{equation}
and the pitch moment $T^b_y$ is computed analogously using the center-of-pressure position
\[
\bm{x}^w_c = [0,\;R/2,\;0]^T.
\]

\subsection{Virtual environment: nonlinear time-invariant (NLTI) dynamics}\label{sec_NLTI}

The virtual environment used in the reinforcement twinning framework is based on a nonlinear time-invariant (NLTI) approximation of the drone dynamics. Compared to the NLTV formulation of the real environment, NLTI models provide a computationally efficient representation that is better suited for repeated policy optimization and real-time implementation. In particular, such reduced-order models offer the potential for on-board deployment on embedded systems, while also enabling the use of classical stability and sensitivity analysis tools.

A common approach to derive NLTI models for flapping-wing systems relies on the averaging theorem \citep{Khalil2002,Schenato2003}. Starting from the NLTV system in equation~\eqref{eq_fs}, a change of time variable $\tau = 2\pi f_c t$ yields the canonical form
\begin{equation}
    \frac{d \bm{s}^b}{d \tau}
    =
    \epsilon \left[
    \bm{g}_1(\bm{s}^b,\bm{a},\tau)
    +
    \bm{g}_2(\bm{s}^b)
    \right],
\end{equation}
where $\epsilon = 1/(2\pi f_c)$ is a small parameter associated with the fast wingbeat dynamics.

For sufficiently small $\epsilon$, the averaging theorem allows the explicit time dependence to be removed, leading to the approximate NLTI system \citep{Taha2014}
\begin{equation}
\label{eq_sdav}
    \frac{d \overline{\bm{s}}^b}{d t}
    =
    \overline{\bm{g}}_1(\overline{\bm{s}}^b,\bm{a},\bm{w}_p)
    +
    \bm{g}_2(\overline{\bm{s}}^b),
\end{equation}
where $\overline{\bm{s}}^b$ denotes the cycle-averaged state and
\begin{equation}
\overline{\bm{g}}_1(\overline{\bm{s}}^b,\bm{a})
=
f_c \int_0^{1/f_c}
\bm{g}_1(\overline{\bm{s}}^b,\bm{a},\tau)\,d\tau    
\end{equation}

is the cycle-averaged aerodynamic contribution.

In practice, evaluating this integral analytically is not trivial due to the nonlinear dependence of the aerodynamic forces on both the state and the control inputs. Existing approaches therefore rely on additional simplifications, such as neglecting state dependence \citep{Schenato2003,Deng2006,Oppenheimer2011} or truncating higher-order terms \citep{Cheng2011,Taha2014}. While effective in restricted operating conditions, these approximations can lead to limited predictive accuracy across the wide range of trajectories encountered during learning.

To overcome these limitations, the present work adopts a data-driven closure for $\overline{\bm{g}}_1$, in which the cycle-averaged aerodynamic forces are represented as a polynomial function of the state and control variables. Specifically, the closure is written as
\begin{align}
\overline{\bm{g}}_1
    &=
\bm{W}_0 
    \begin{bmatrix}
    A_\phi \\
    \beta \\
    A_{\text{off}}
    \end{bmatrix}
+
\bm{W}_1 
    \begin{bmatrix}
    A_\phi^2 \\
    \beta^2 \\
    A_{\text{off}}^2
    \end{bmatrix}
+
\bm{W}_2
    \begin{bmatrix}
    A_\phi A_{\text{off}} \\
    A_\phi \beta \\
    A_{\text{off}} \beta
    \end{bmatrix}
\notag \\
&\quad
+
\bm{W}_3
    \begin{bmatrix}
    \dot{\overline{x}} \\
    \dot{\overline{z}} \\
    \dot{\overline{\theta}}
    \end{bmatrix}
+
\bm{W}_4
    \begin{bmatrix}
    \dot{\overline{x}}\dot{\overline{z}} \\
    \dot{\overline{x}}\dot{\overline{\theta}} \\
    \dot{\overline{\theta}}\dot{\overline{z}}
    \end{bmatrix}
\notag \\
&\quad
+
\bm{W}_5
    \begin{bmatrix}
    \dot{\overline{x}}^2 \\
    \dot{\overline{z}}^2 \\
    \dot{\overline{\theta}}^2
    \end{bmatrix}
+
\bm{W}_6
    \begin{bmatrix}
    A_\phi \dot{\overline{x}} \\
    \beta \dot{\overline{x}} \\
    A_{\text{off}} \dot{\overline{x}}
    \end{bmatrix}
\notag \\
&\quad
+
\bm{W}_7
    \begin{bmatrix}
    A_\phi \dot{\overline{z}} \\
    \beta \dot{\overline{z}} \\
    A_{\text{off}} \dot{\overline{z}}
    \end{bmatrix}
+
\bm{W}_8
    \begin{bmatrix}
    A_\phi \dot{\overline{\theta}} \\
    \beta \dot{\overline{\theta}} \\
    A_{\text{off}} \dot{\overline{\theta}}
    \end{bmatrix},
\label{eq_Fav}
\end{align}
where $\bm{W}_0,\ldots,\bm{W}_8 \in \mathbb{R}^{3\times3}$ are the closure matrices. The full model therefore involves $n_{w_p}=81$ parameters, which are identified online through the adjoint-based data-assimilation procedure described in Section~\ref{sec_adjoint}.

This polynomial representation provides a flexible yet computationally efficient surrogate of the cycle-averaged aerodynamic forces. In contrast to classical reduced-order models, it allows the dependence on both control inputs and state variables to be retained, while remaining compatible with fast gradient-based optimization.

To assess the trade-off between model complexity and predictive capability, several reduced versions of the closure are considered. These are labelled from Model~1 (M1) to Model~5 (M5), with increasing levels of complexity. The full formulation in equation~\eqref{eq_Fav} corresponds to M5. The simplest model (M1) neglects all state-dependent terms, i.e.
\[
\bm{W}_3=\bm{W}_4=\bm{W}_5=\bm{W}_6=\bm{W}_7=\bm{W}_8=0.
\]
Model~2 (M2) corresponds to the classical linear time-invariant (LTI) formulation commonly used in hovering-flight studies \citep{Taha2012,Orlowski2012}, retaining only the linear control and linear state terms. Model~3 (M3) extends M2 by including quadratic control terms, while Model~4 (M4) further incorporates second-order state contributions. These variants are compared in Section~\ref{sec_off} to evaluate their impact on control performance and model accuracy.

\section{Results}\label{sec_results}

This section is organized into four subsections. Section~\ref{sec_off} evaluates the NLTI surrogate models introduced in Section~\ref{sec_NLTI} and identifies the best compromise between predictive accuracy and model complexity. The selected model is then used as the adaptive surrogate within the reinforcement twinning (RT) framework.

The following three subsections assess the control performance of RT under different initializations of the closure-law parameters $\bm{w}_p$. In the first scenario (Section~\ref{C1}), RT starts from an offline-calibrated model, representing the case where a sufficiently rich dataset is available prior to deployment. In the second scenario (Section~\ref{C2}), both the model and the policy are initialized randomly, corresponding to a fully online learning setting with no prior information. In the third scenario (Section~\ref{C3}), the initial model is biased, mimicking situations in which the FWMAV dynamics have changed during operation or the available pre-training data are no longer fully representative. In all three cases, the objective is to evaluate how efficiently the RT framework can learn an effective control policy while adapting the surrogate model online.

\begin{figure*}
    \centering
    \begin{subfigure}[t]{0.6\linewidth}
        \centering
        \includegraphics[width=0.7\linewidth]{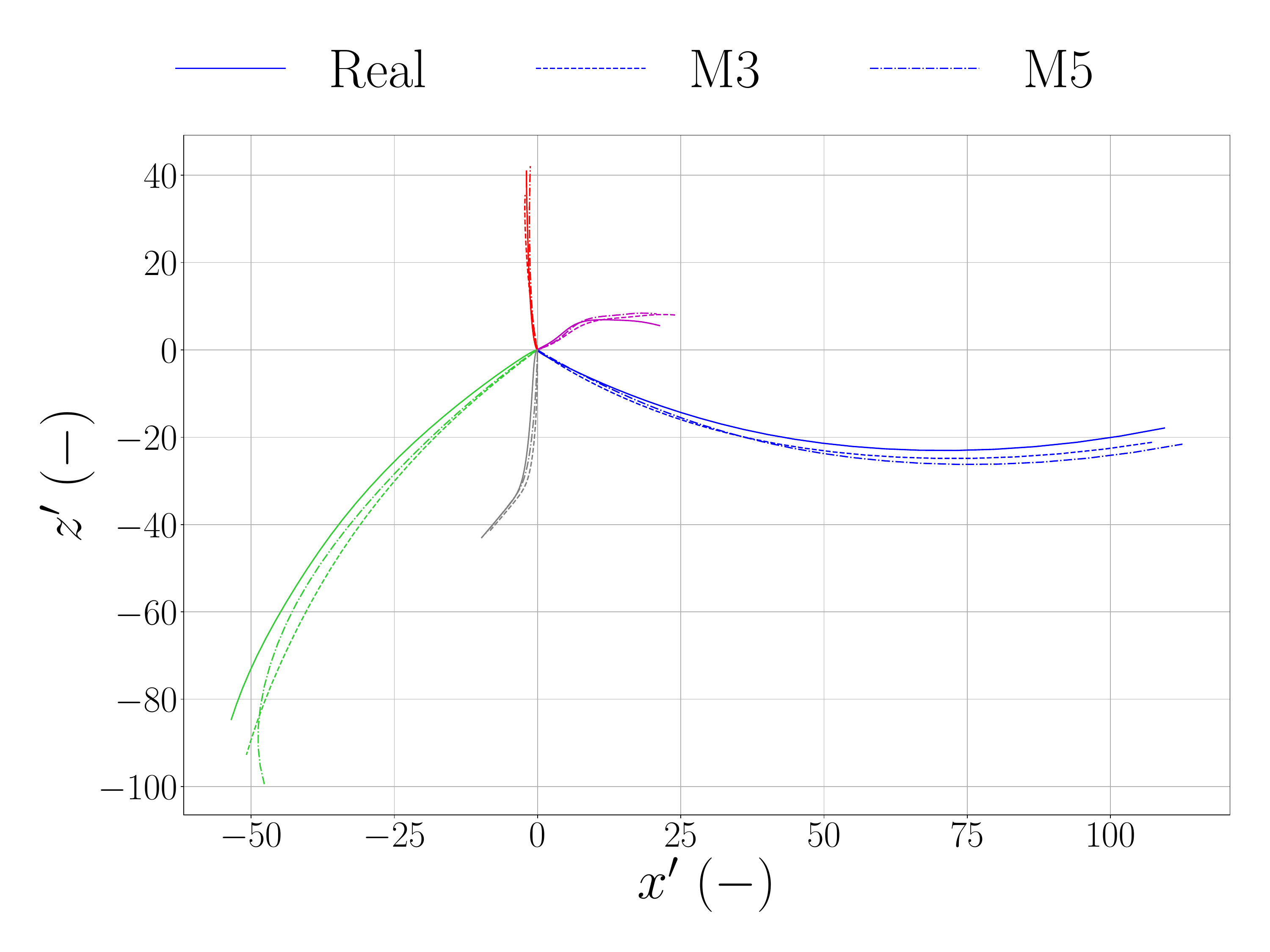}
        \caption{}
         \label{fig_assim_traj_a}
    \end{subfigure}
    \begin{subfigure}[t]{0.6\linewidth}
        \centering
        \includegraphics[width=\linewidth]{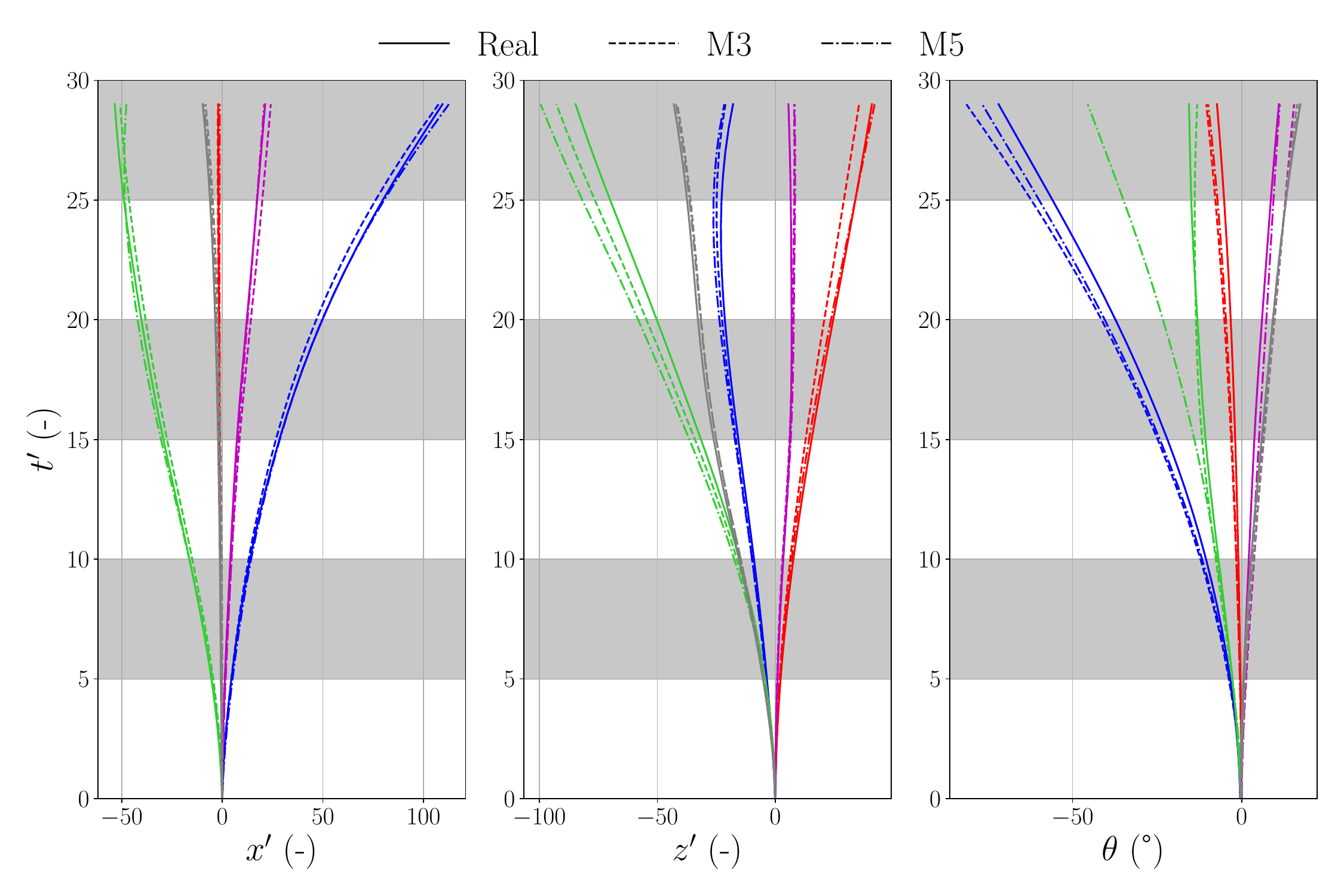}
        \caption{}
        \label{fig_assim_traj_b}
    \end{subfigure}
    \caption{(a) Comparison of open-loop trajectories generated by the real environment and the virtual environment (Model 3 (M3) and Model 5 (M5)) using the testing dataset, and (b) detailed view of the $x$, $z$, and $\theta$ states of the trajectories over time.}
    \label{fig_assim_traj}
\end{figure*}

\begin{figure*}
    \centering
    \includegraphics[width=0.7\linewidth]{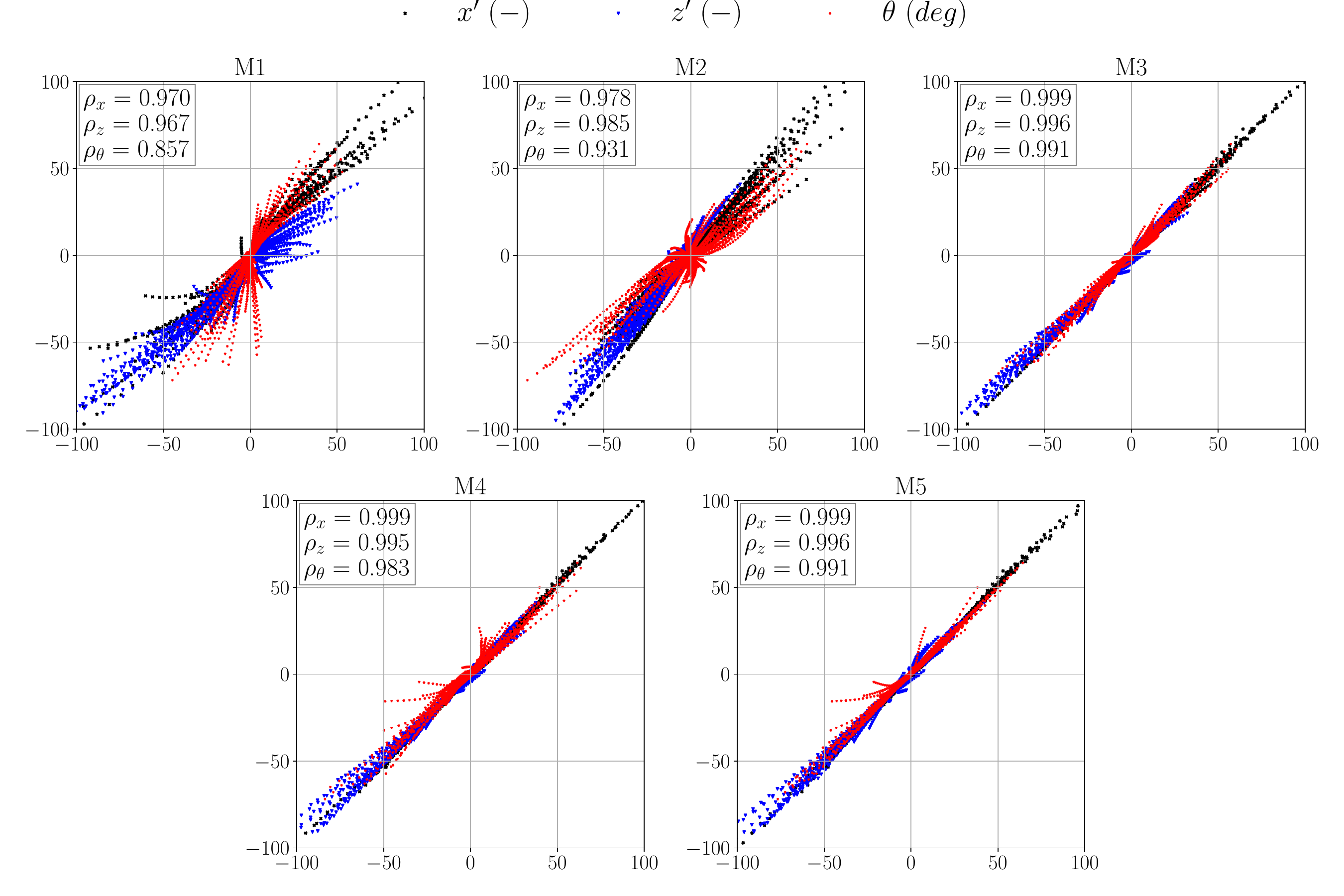}
    \caption{Scatter plot of the trajectories computed with the real and the virtual environment for five closure laws assimilated from the real trajectories.}
    \label{fig_pearson}
\end{figure*}

\subsection{Model selection}\label{sec_off}

To evaluate the NLTI surrogates introduced in equation~\eqref{eq_Fav}, a database of $n_{tr}=100$ trajectories is generated using the real environment and random control actions. These actions are sampled from a stationary Gaussian process in time, with a randomly selected mean within the admissible action space and a covariance length scale chosen to ensure smooth variations of the control inputs over each episode. All trajectories start from $\check{\bm{s}}=\bm{0}$ and are cycle-averaged to ensure consistency with the virtual environment. Twenty of the 100 trajectories are used to identify the NLTI models through the adjoint-based procedure described in Section~\ref{sec_adjoint}, while the remaining 80 are reserved for testing.

Figure~\ref{fig_assim_traj_a} shows five representative test trajectories in terms of the dimensionless states $x'=x/R$ and $z'=z/R$, where $R$ denotes the wing span. The trajectories predicted by the real NLTV environment are compared with those obtained using the M3 and M5 NLTI surrogates. Both models reproduce the real trajectories closely. Figure~\ref{fig_assim_traj_b} provides a more detailed comparison of the three degrees of freedom as a function of the dimensionless time $t'=tf_c$. These results indicate that M3, which contains 39 trainable parameters, achieves a predictive performance comparable to that of M5, which contains 81 parameters. The discrepancy between the NLTI and NLTV trajectories becomes more visible only for long distances, typically over distances of 5 to 6 m.

A systematic comparison of all five surrogate models is reported in Figure~\ref{fig_pearson}, which shows the integrated testing error over the full trajectories together with the corresponding Pearson correlation coefficients. Model~1 exhibits the largest dispersion, highlighting the importance of state feedback in the force formulation. Model~2, which introduces linear state dependence, improves the agreement between real and virtual trajectories, although noticeable dispersion remains. This indicates that higher-order terms are required to capture the range of flight regimes encountered in the dataset. Model~3, which adds quadratic control terms, markedly improves the predictions. In particular, the longitudinal, vertical, and pitching positions all exhibit a clear linear trend, with Pearson correlation coefficients above 0.99. Including second-order state terms (i.e. $\bm{W}_4$ and $\bm{W}_5$ in equation~\eqref{eq_Fav}) in M4, or additional state--control cross terms (i.e. $\bm{W}_6$, $\bm{W}_7$, and $\bm{W}_8$) in M5, does not yield a significant further improvement for the trajectory lengths considered here. For the remainder of the paper, Model~3 is therefore retained as the best compromise between accuracy and complexity.

\subsection{Policy training with offline model calibration}\label{C1}

 This section evaluates the performance of the RT algorithm in training the control policy (equation \ref{eq_pi}) using M3 as the closure law for the virtual environment. The model is pre-calibrated offline as in Section \ref{sec_off}, ensuring state predictions similar to the real environment. It is then assumed that the real-to-virtual environment trust ratio satisfies $RVET_l<RVET<RVET_u$ from the outset of policy training (see Figure \ref{fig_diag2}). Table \ref{tab_trainParameters} lists the key hyperparameters used for the training. The parameters were selected through a trial-and-error process, starting from the reference values established in \cite{Schena2023}. To quantify the benefit of combining model-based and model-free learning, RT is compared against a standard stand-alone DDPG implementation, corresponding to Steps 1 and 2 of Algorithm~\ref{alg:main}.

\begin{table*}[htbp]
\centering

\begin{tabularx}{\textwidth}{|>{\centering\arraybackslash}X|>{\centering\arraybackslash}X|>{\centering\arraybackslash}X|>{\centering\arraybackslash}X|>{\centering\arraybackslash}X|}
    \hline
    \multicolumn{5}{|c|}{\textbf{General Parameters}} \\ \hline
    $T_{0}$ (s) & $\Delta t$ (s) & $N_{c}$ & $N_{ep}$ & $N_{s}$ \\ \hline
    1.5 & 0.05 & 30 & 250 & 25 \\ \hline
\end{tabularx}

\vspace{5pt} 

\begin{tabularx}{\textwidth}{|>{\centering\arraybackslash}X|>{\centering\arraybackslash}X|>{\centering\arraybackslash}X|>{\centering\arraybackslash}X|>{\centering\arraybackslash}X|>{\centering\arraybackslash}X|>{\centering\arraybackslash}X|}
    \hline
    \multicolumn{7}{|c|}{\textbf{Model Parameters}} \\ \hline
    \multicolumn{3}{|c|}{\textbf{Model-Free}} & \multicolumn{4}{c|}{\textbf{Model-Based}} \\ \hline
    $n_{R}$ & $n_a$ & $n_q$ & $n_p$ & $n_e$ & $n_G$ & $n_{mb}$ \\ \hline
    2000 & 30 & 30 & 39 & 5 & 50 & 30 \\ \hline
\end{tabularx}

\vspace{5pt} 

\begin{tabularx}{\textwidth}{|>{\centering\arraybackslash}X|>{\centering\arraybackslash}X|>{\centering\arraybackslash}X|>{\centering\arraybackslash}X|>{\centering\arraybackslash}X|>{\centering\arraybackslash}X|>{\centering\arraybackslash}X|>{\centering\arraybackslash}X|}
    \hline
    \multicolumn{5}{|c|}{\textbf{Referee Parameters}} \\ \hline
     $T_{\mathcal{J}_1}$ & $T_{\mathcal{J}_2}$ & $T_w$ & $RVET_l$ & $RVET_u$ \\ \hline
     1 & 0.8 & 0.05 & 0 & 40 \\ \hline
\end{tabularx}

\caption{Hyperparameters used by the reinforcement twinning algorithm.}
\label{tab_trainParameters}
\end{table*}

\begin{figure*}[ht]
\centering
\begin{subfigure}{0.48\textwidth}
    \centering
    \includegraphics[width=\textwidth]{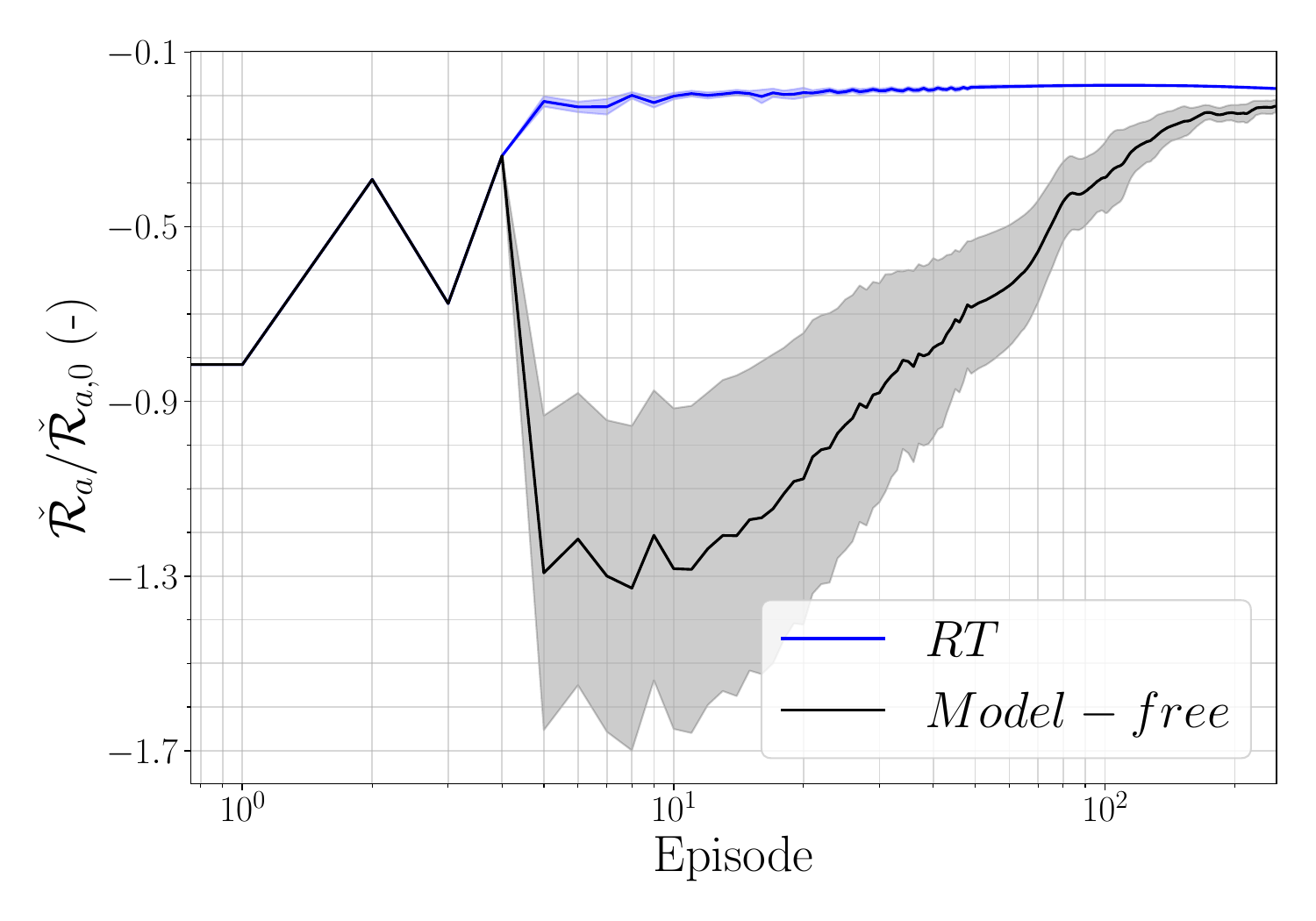}
    \caption{}
    \label{fig_learningControl_a}
\end{subfigure}%
\hfill
\begin{subfigure}{0.48\textwidth}
    \centering
    \includegraphics[width=\textwidth]{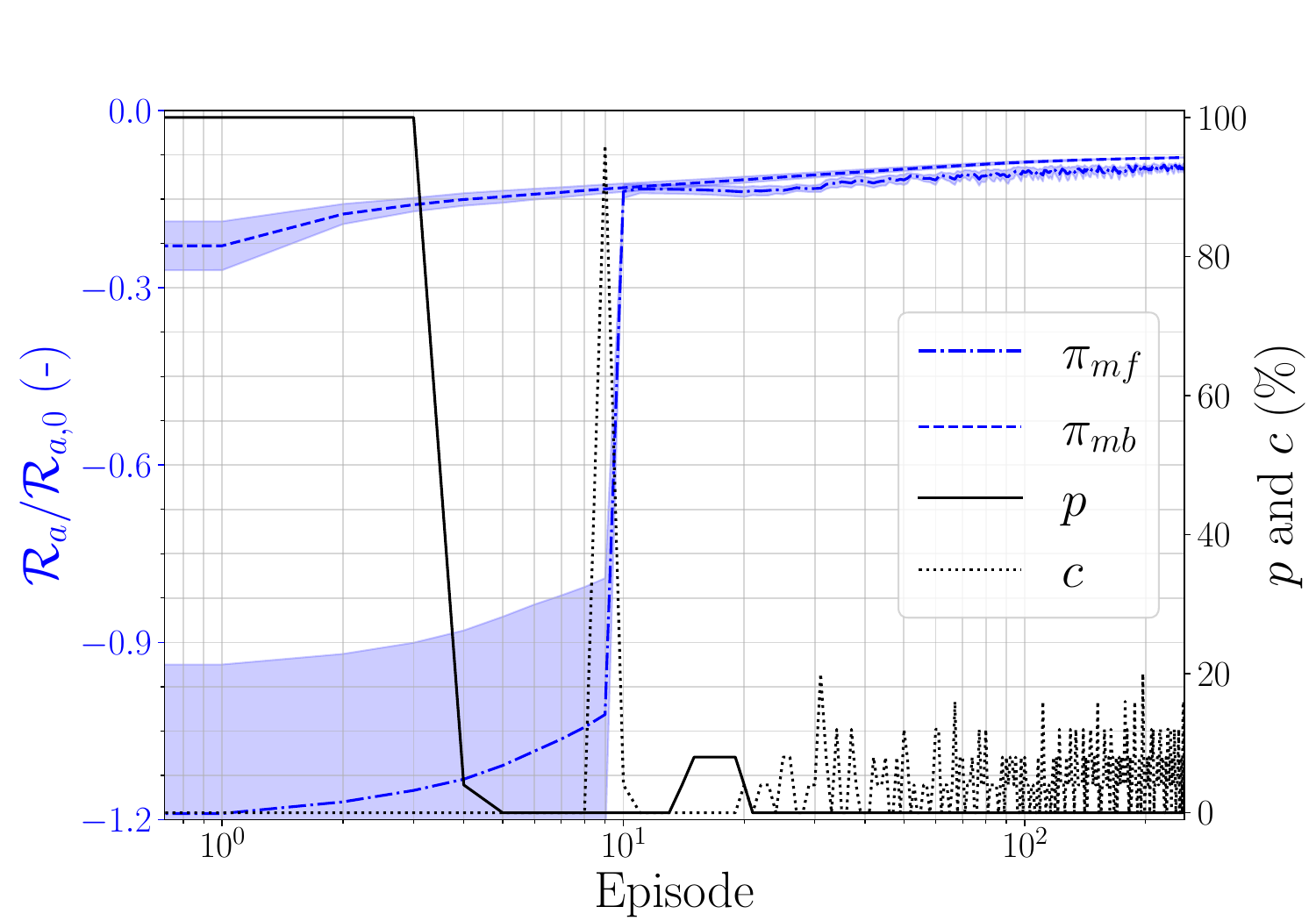}
    \caption{}
    \label{fig_learningControl_b}
\end{subfigure}
\caption{Comparison of cumulative rewards for (a) RT and DDPG training and (b) model-based and model-free policies, along with policy type and weight cloning signals during RT training.}
\label{fig_learningControl}
\end{figure*}

Figure~\ref{fig_learningControl_a} reports the cumulative reward (equation~\ref{eq_Ra}) over $N_{ep}=250$ training episodes. The curves show the mean and 95\% confidence interval over $N_s=25$ runs with randomly initialized policy weights. During the first five episodes, control actions are sampled from the trajectory database to populate the replay buffer and promote initial exploration. This bootstrapping phase accelerates the convergence of the model-free branch in both RT and stand-alone DDPG. Once this phase ends, the RT reward rises rapidly and converges to the highest value with negligible variance, whereas the stand-alone model-free policy improves more gradually and converges to a lower mean reward with larger variability.

A more detailed view of RT training is provided in Figure~\ref{fig_learningControl_b}, which reports the cumulative reward of the model-based (dashed line) and model-free (dash-dotted line) policies, both evaluated in the virtual environment. The policy type signal $p$ is also shown indicating the live policy percentage which are using the model-free policy over the $N_s$ training runs. Similarly, the weight-cloning signal $c$ shows the percentage of policy cloning operations over the $N_s$ parallel runs.

By default, the live policy is initially model-free. Its cumulative reward remains consistently higher than that of the model-based policy, after which a policy switch occurs and the model-based policy becomes live. Over the next episodes, the model-based policy consistently outperforms the model-free one, triggering the weight-cloning mechanism and copying $\bm{w}_{a,mb}$ into $\bm{w}_{a,mf}$. As a result, the cumulative reward of the model-free policy increases sharply around episode 9. The remaining training is then conducted mostly under model-based control, while the two reward curves progressively converge and multiple cloning events are observed. In this favorable setting, where the virtual environment is already well aligned with the real system, the model-based branch effectively drives policy learning and acts as a safeguard for model-free exploration.

\begin{figure*}[ht]
\centering

\begin{subfigure}{0.85\textwidth}
    \centering
    \includegraphics[width=\textwidth]{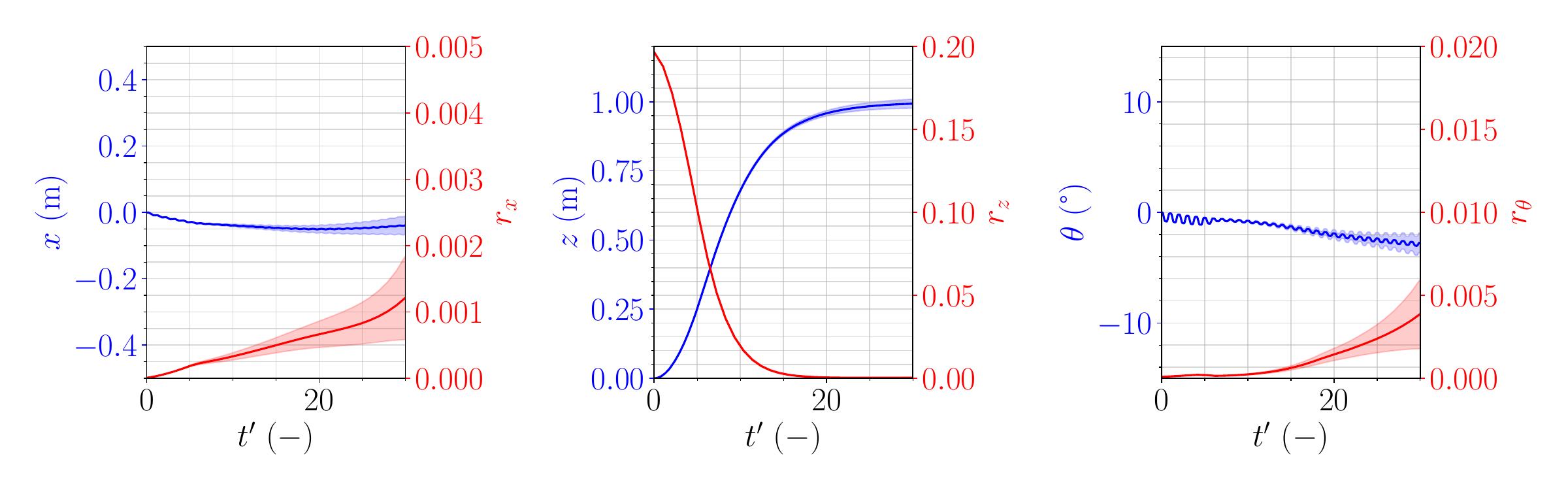}
    \caption{}
    \label{fig_case1_x}
\end{subfigure}

\begin{subfigure}{0.85\textwidth}
    \centering
    \includegraphics[width=\textwidth]{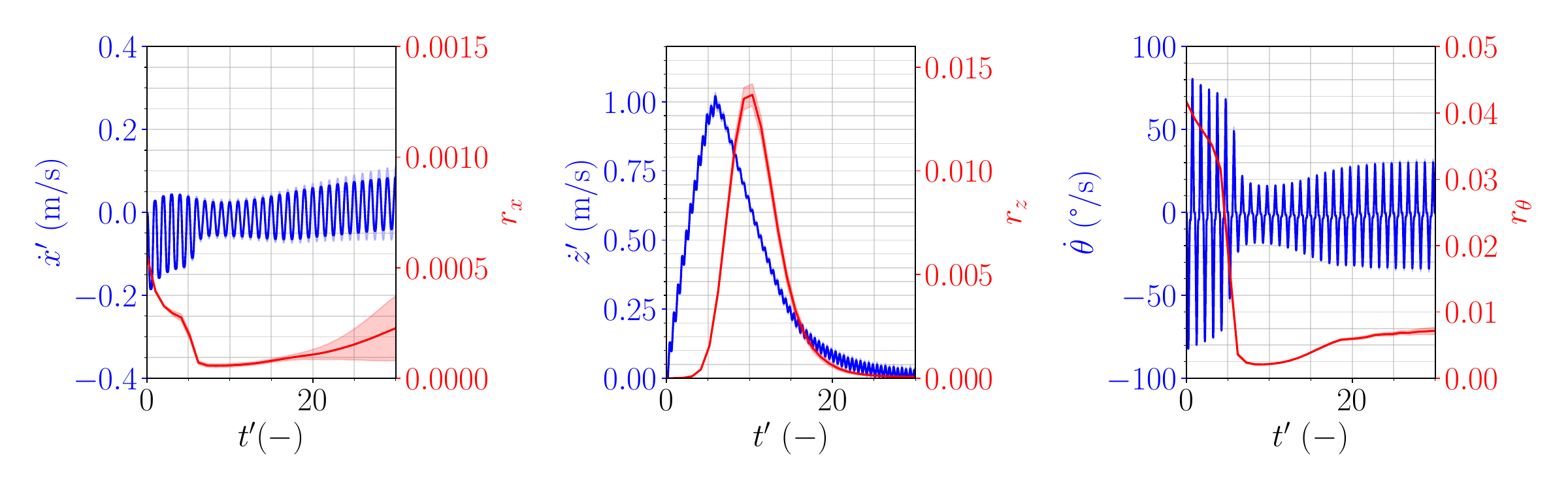}
    \caption{}
    \label{fig_case1_xd}
\end{subfigure}

\begin{subfigure}{0.85\textwidth}
    \centering
    \includegraphics[width=\textwidth]{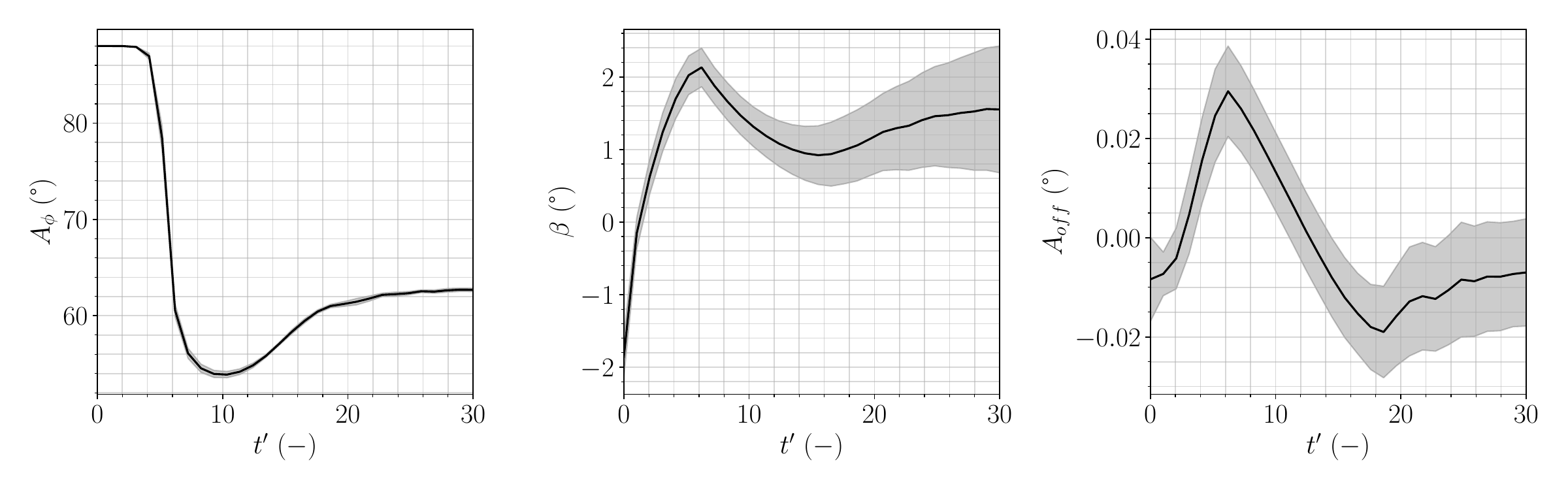}
    \caption{}
    \label{fig_case1_a}
\end{subfigure}

\caption{Vertical ascent and hover trajectories shown with (a) the drone positions, (b) the drone velocities resulting from a RT trained policy, and (c) the control actions.}
\label{fig_case1_all}
\end{figure*}

In this scenario, the model-free branch contributes little to the improvement of the model-based policy, but such support is not required: the model-based branch does not encounter significant local minima and transfers effectively from the virtual to the real environment.

The resulting control performance is illustrated in Figure~\ref{fig_case1_all}. Figure~\ref{fig_case1_x} shows the time evolution of the dimensionless positions $x'$ and $z'$ together with the pitch angle $\theta$, while Figure~\ref{fig_case1_xd} reports the corresponding velocities. In both panels, the secondary axis indicates the contribution of each degree of freedom to the cycle reward in equation~\eqref{eq_rk}. The curves represent the mean and 95\% confidence interval over the $N_s$ training runs. The high-frequency oscillations reflect the periodic aerodynamic forcing induced by the wingbeat motion.

Figures~\ref{fig_case1_x} and \ref{fig_case1_xd} show that the drone initially accelerates upward over the first few flapping cycles, reaches a peak vertical velocity, and then decelerates before stabilizing near $z=1$ m ($z'=20$). This behavior is directly linked to the flapping amplitude $A_\phi$, which starts near its maximum, decreases as the target is approached, and eventually settles to a value for which lift approximately balances weight. Through the coupled dynamics of equation~\eqref{eq_instSystem}, the flapping amplitude also affects the longitudinal and pitching motions, which exhibit small but nonzero deviations. These secondary dynamics are compensated by modest adjustments of the stroke-plane angle $\beta$ and mean flapping angle $A_{\text{off}}$. The reward decomposition further shows that the vertical-position error dominates the reward during roughly the first 20 cycles, whereas the pitch contribution becomes more important later owing to a slight heading drift.

The markedly different variances of the control signals in Figure~\ref{fig_case1_a} also provide insight into their relative importance. In particular, the flapping amplitude $A_\phi$ has a much stronger influence on task success than either the stroke-plane angle $\beta$ or the mean offset $A_{\text{off}}$. This suggests that alternative reward shaping could further improve performance. For instance, increasing the weight associated with vertical-position tracking might accelerate convergence, although at the cost of slightly larger drift in the remaining degrees of freedom.

\subsection{Policy training with online twinning}\label{C2}

The previous case assumes that a sufficiently accurate surrogate model is available before training begins. In practice, however, collecting such a dataset may be difficult or unsafe, especially when the FWMAV dynamics are initially unknown. Fully random control actions may produce damaging trajectories, whereas overly restricted excitation may fail to generate sufficiently informative data. In addition, sequentially identifying the model and then training the policy can be sample-inefficient, since even a partially calibrated model may already provide useful guidance to both the model-based and model-free branches. This section therefore considers the more demanding setting in which both the surrogate model and the control policy are trained online from randomly initialized $\bm{w}_p$ and $\bm{w}_a$. The training is conducted over $N_s$ runs using the same hyperparameters as in the previous test case (Table \ref{tab_trainParameters}).

Four strategies are compared:  
(1) a purely model-free baseline, corresponding to Steps 1 and 2 of Algorithm~\ref{alg:main}; (2) a purely model-based approach, corresponding to Steps 1 and 3--6; and (3)--(4) two full RT implementations that differ only in the update criterion used for the assimilation buffer $\mathcal{D}_A$, namely maximum variance and minimum error as introduced in Section~\ref{eq_generalPrinciple} As for the cases in Section \ref{C1}, actions are sampled from a Gaussian process for the first $n_e=5$ episodes for all methods.  This allows populating the assimilation and replay buffers ($\mathcal{D}_A$ and $\mathcal{D}_{R}$) with demonstration trajectories, accelerating the training of the model, the critic network, and the model-free policy. Once the assimilation buffer is full, the policy begins interacting with the environment, and the assimilation buffer is updated following either the maximum-variance or the minimum-error criteria.

\begin{figure*}[ht]
    \begin{subfigure}{0.65\textwidth}
        \includegraphics[width=\textwidth]{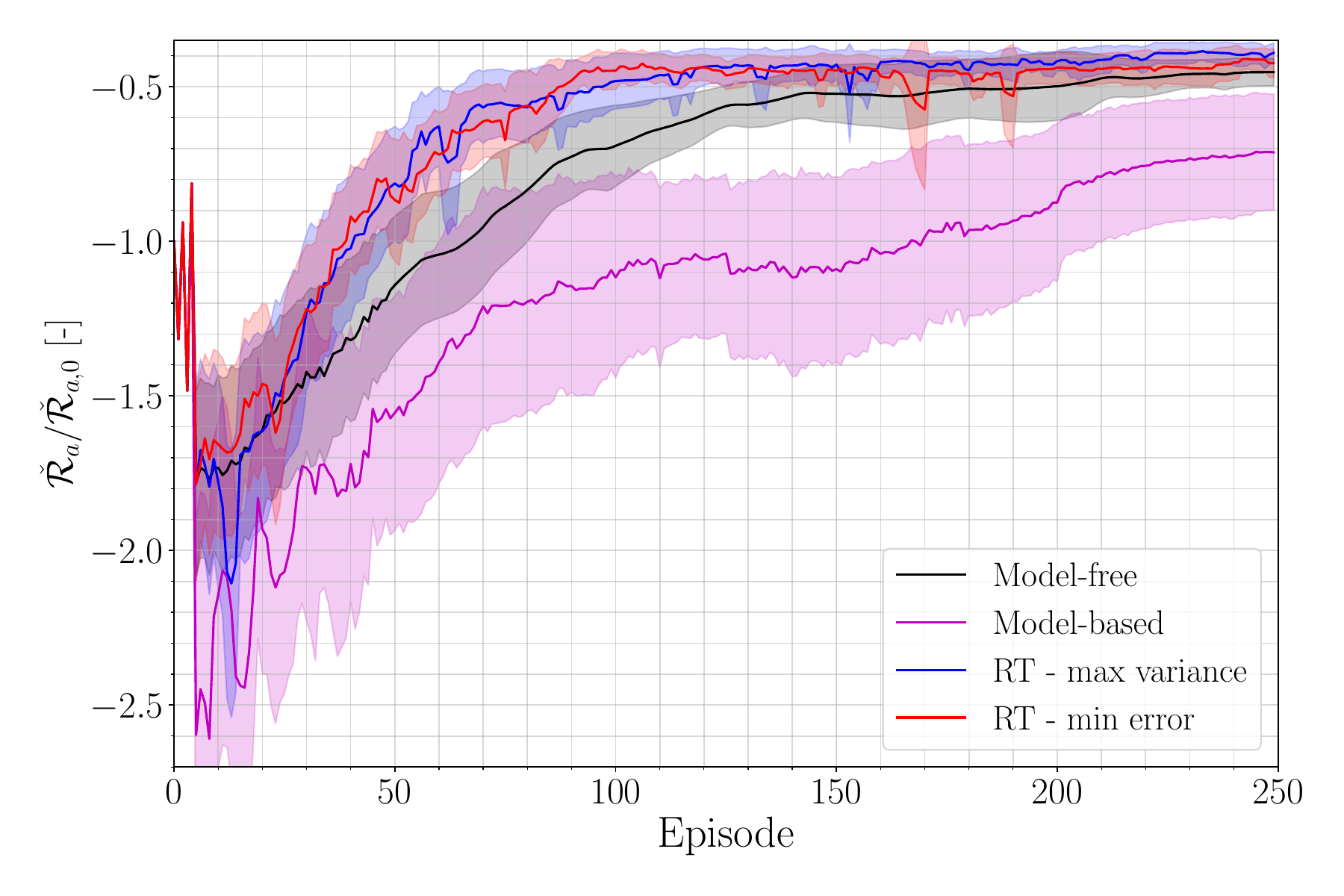}
        \caption{}
        \label{fig_case2_learningControl_a}
    \end{subfigure}
    \centering
    \begin{subfigure}{0.42\textwidth}
        \includegraphics[width=\textwidth]{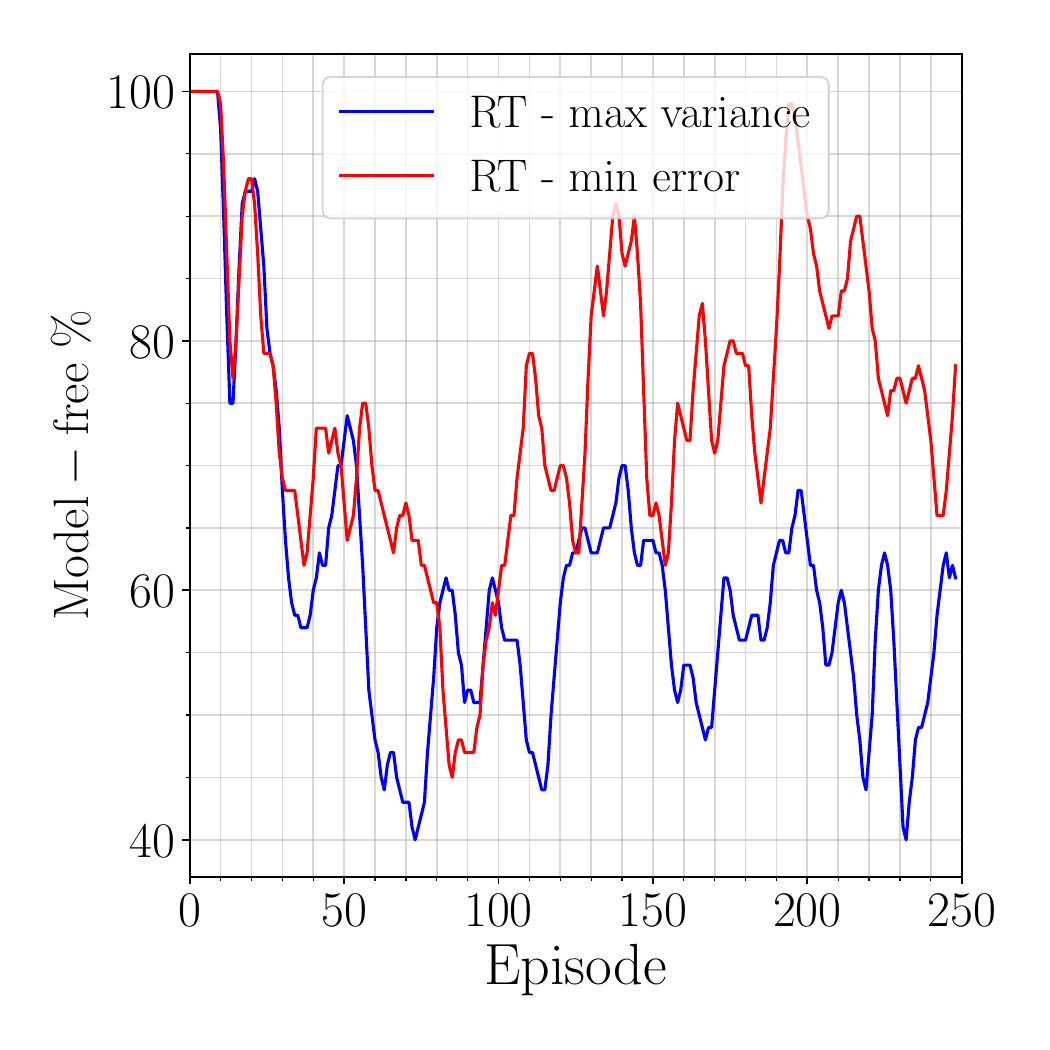}
        \caption{}
        \label{fig_case2_learningControl_b1}
    \end{subfigure}
    \begin{subfigure}{0.42\textwidth}
        \includegraphics[width=\textwidth]{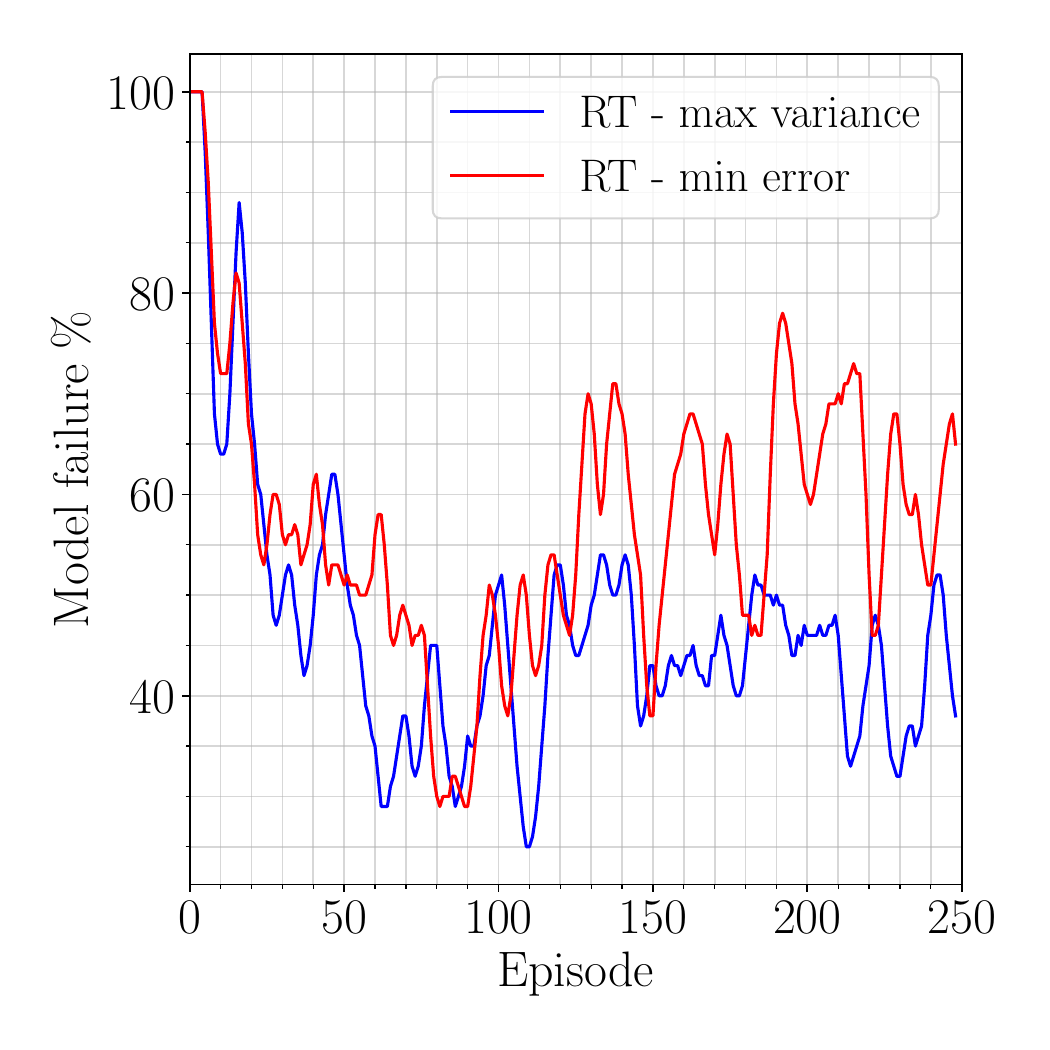}
        \caption{}
        \label{fig_case2_learningControl_b2}
    \end{subfigure}
    \caption{(a) Comparison of cumulative rewards among the model-free approach, the model-based approach, and the RT approach, using two different saving strategies for the assimilation buffer. (b) Comparison of the model-free percentage and (c) model-failure percentage for the two strategies.}
    \label{fig_case2_learningControl}
\end{figure*}

 Figure \ref{fig_case2_learningControl_a} shows the evolution of the cumulative reward for the four approaches. In terms of sample efficiency, the fully model-based strategy performs worst: it converges most slowly, reaches the lowest mean reward, and exhibits the largest variance. This behavior reflects its complete reliance on the surrogate model, whose parameters $\bm{w}_p$ must be identified while the policy is being optimized. Since these parameters are initialized randomly, the early policy updates are based on a poor model, which induces policy bias: improvements observed in the virtual environment do not reliably translate to the real system. Interestingly, this mismatch also causes the live policy to explore widely separated regions of the action space. Although this degrades control performance, it also enriches the assimilation buffer and indirectly improves model calibration.
  
Under such poor initial modeling conditions, the purely model-free baseline clearly outperforms the purely model-based approach in terms of convergence rate, variance, and final cumulative reward. This places the present test case at the opposite end of the spectrum from Section~\ref{C1}, and therefore provides a relevant benchmark for assessing the benefit of hybrid learning. Both RT variants outperform the purely model-free and purely model-based baselines, achieving faster convergence and higher final rewards. In both cases, RT overtakes the purely model-free approach around episode 30. 

This transition can be interpreted with the help of Figures~\ref{fig_case2_learningControl_b1} and \ref{fig_case2_learningControl_b2}. Figure~\ref{fig_case2_learningControl_b1} reports the average percentage of runs in which the live policy is model-free, whereas Figure~\ref{fig_case2_learningControl_b2} shows the percentage of runs for which the surrogate is inconsistent with the real environment according to the RVET criterion (equation~\eqref{eq_MET}). Around episode 30, the model-failure percentage decreases substantially, indicating that data assimilation has improved the consistency of the virtual environment. Once this occurs, the model-based branch becomes reliable enough to contribute effectively to policy optimization. Owing to its higher sample efficiency, the model-based policy then rapidly improves and overtakes the model-free policy, as reflected by the drop in the model-free percentage. As training continues, however, the live policy reaches new regions of the state space where the surrogate is again less accurate, causing a temporary return to model-free control. This alternating pattern between short model-based phases and longer model-free phases persists until the end of the training.

For the maximum-variance assimilation strategy, the mean model-failure percentage decreases during training, which leads to a progressively larger contribution of model-based actions in the real environment. By contrast, for the minimum-error strategy, the model-failure percentage increases over time, and the live policy remains predominantly model-free. In other words, the model guides learning only as long as it can maintain sufficient fidelity in the visited regions of the state space. Despite this limitation, both RT variants improve on the purely model-free baseline.

Another important observation is that RT produces significantly smaller variance across runs than the purely model-free strategy, indicating a more structured exploration of the state space. When the model-free branch diverges, the model-based branch often performs better and is promoted to live policy, thereby redirecting exploration toward more promising control regions. Conversely, when the surrogate becomes unreliable, the model-free branch takes over while still benefiting from the transitions previously generated by the model-based policy and stored in the replay buffer.

\begin{figure*}[ht]
    \centering
    \begin{subfigure}{0.48\textwidth}
        \includegraphics[width=\textwidth]{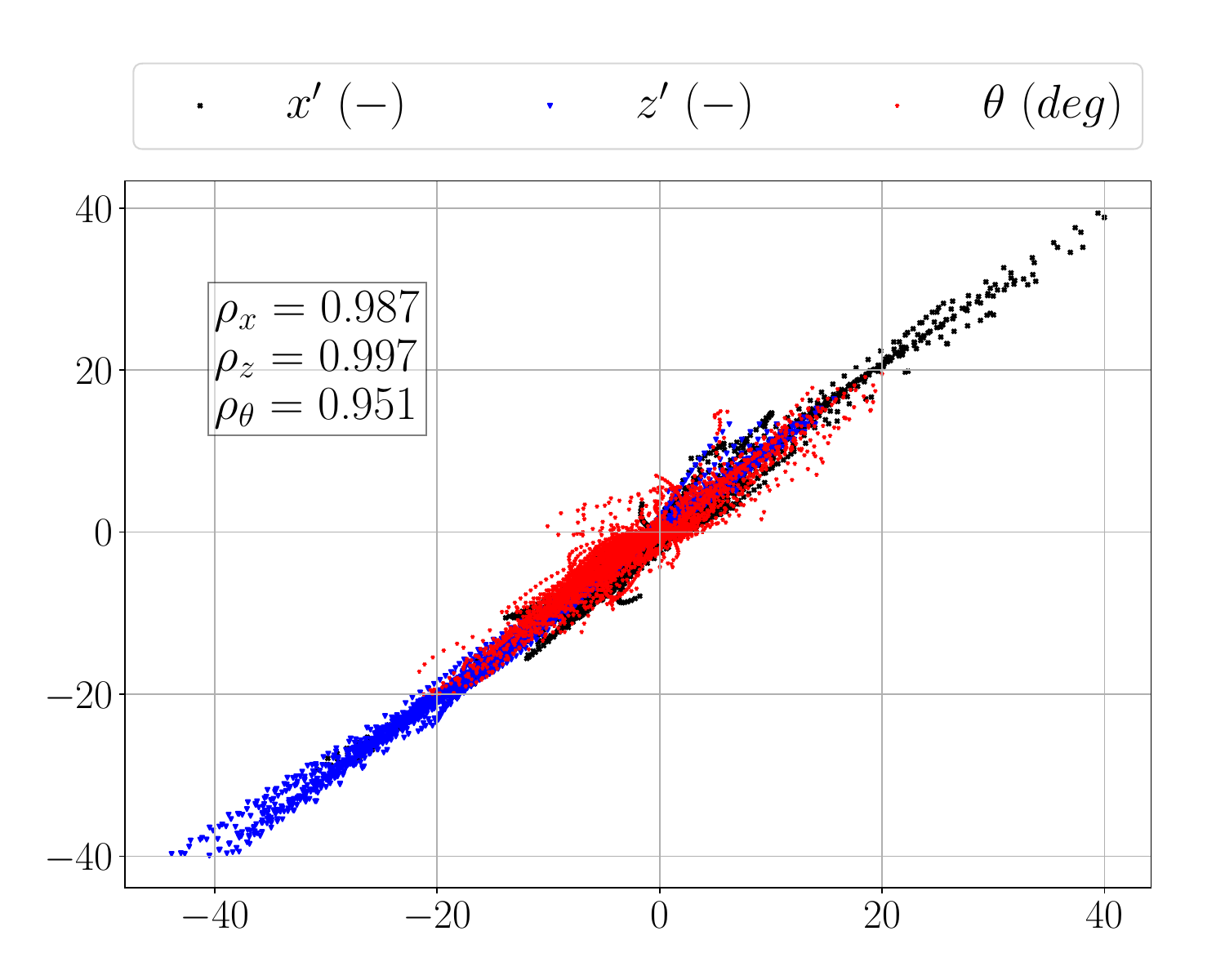}
        \caption{}
        \label{fig_assimBuffer_maxVar}
    \end{subfigure}
    \begin{subfigure}{0.48\textwidth}
        \includegraphics[width=\textwidth]{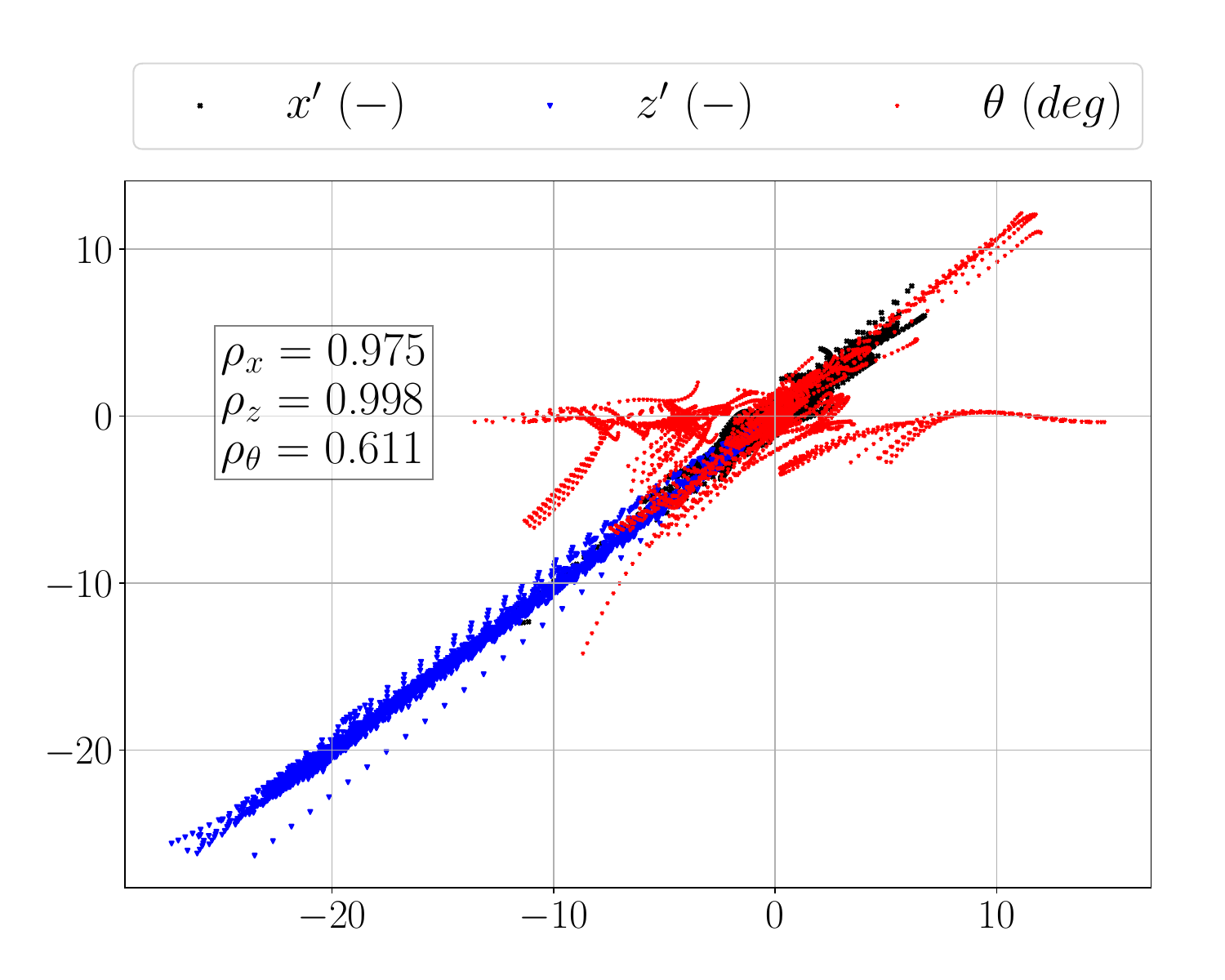}
        \caption{}
        \label{fig_assimBuffer_minErr}
    \end{subfigure}
    \caption{Comparison of the states saved in the assimilation buffer after the RT training following (a) the maximum-variance strategy and (b) the minimum-error strategy.}
    \label{fig_assimBuffer}
\end{figure*}

Figure~\ref{fig_case2_learningControl_a} also shows that maximizing the variance of the assimilation buffer leads to slightly faster convergence than minimizing trajectory-to-target error. This difference is explained by the state distributions reported in Figures~\ref{fig_assimBuffer_maxVar} and \ref{fig_assimBuffer_minErr}. With the minimum-error strategy, the stored trajectories cluster strongly near the target state, which promotes overfitting and reduces the diversity of the training data. By contrast, the maximum-variance strategy preserves a broader coverage of the state space. Nevertheless, both strategies yield successful model assimilation, as reflected by the overall linear trends relating real and virtual states. The lowest Pearson correlation coefficients are observed for the pitching dynamics, indicating that rotational motions remain the most difficult to capture, especially under the minimum-error strategy where the pitch angle remains concentrated near zero. Although additional weighting could be introduced in equation~\eqref{eq_Jp} to reduce model failure further, the current level of accuracy already provides useful guidance to policy learning.

\begin{figure*}[ht]
    \centering
    \begin{subfigure}{0.48\textwidth}
            \includegraphics[width=\textwidth]{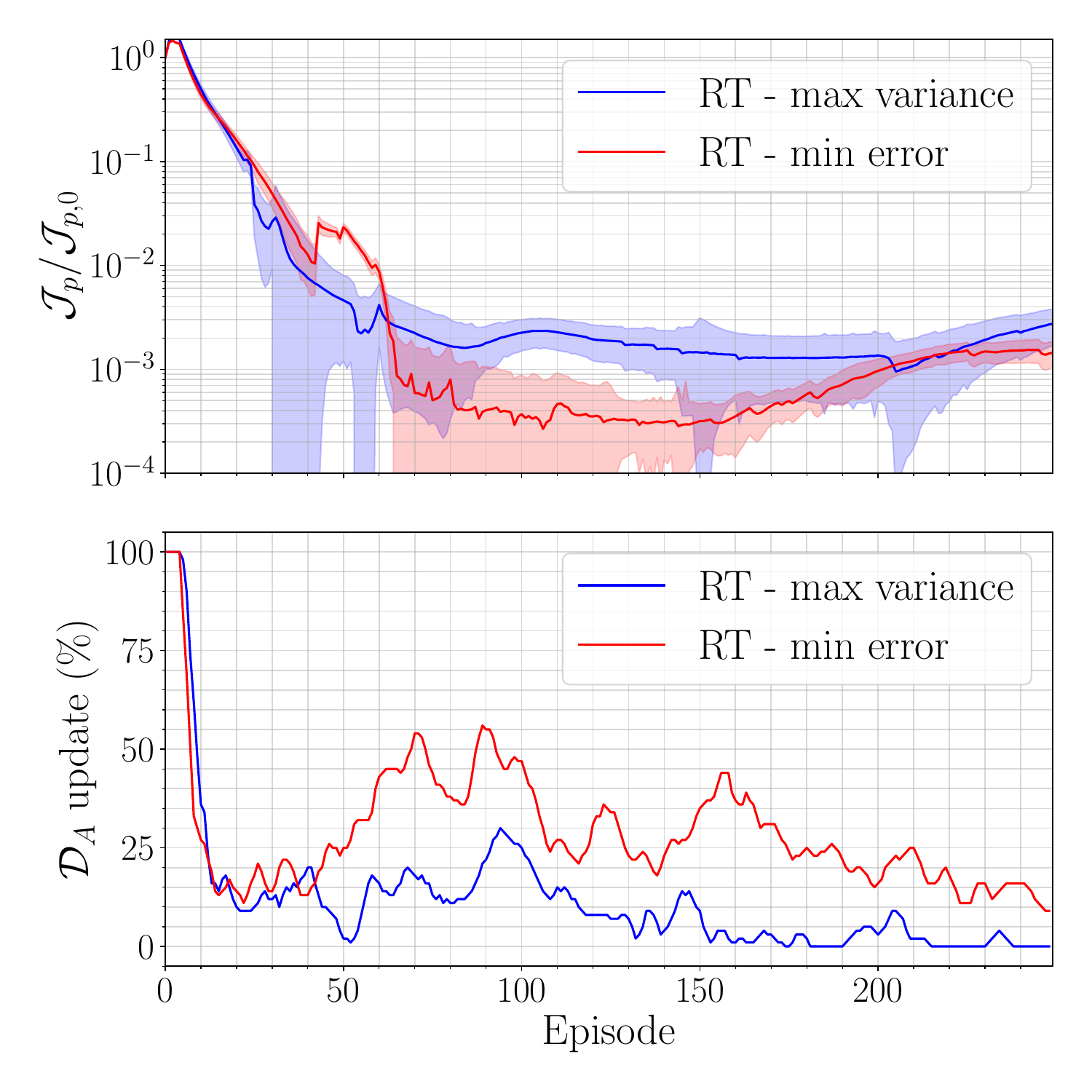}
        \caption{}
        \label{fig_wpjp}
    \end{subfigure}
    \begin{subfigure}{0.48\textwidth}
        \includegraphics[width=\textwidth]{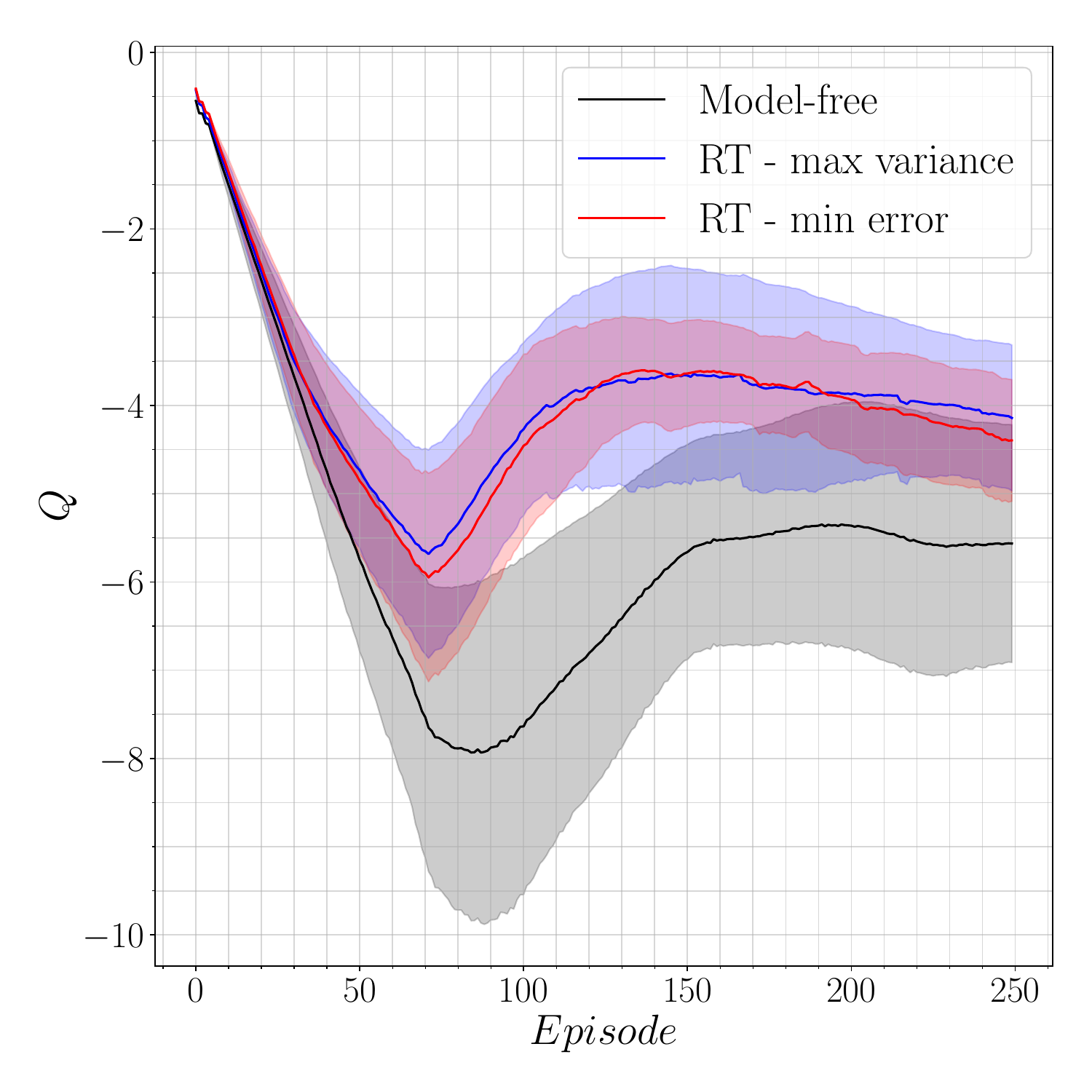}
        \caption{}
        \label{fig_q}
    \end{subfigure}
    \caption{(a) Comparison of the assimilation cost for the two RT training (top) and the update percentage in the assimilation buffer (bottom). (b) Comparison of the Q-value predicted by the critic network during the RT and model-free training.}
    \label{fig_jpandQ}
\end{figure*}

Finally, Figure~\ref{fig_q} reports the Q-value averaged over the mini-batches sampled from the replay buffer during training. An initial overshoot is observed, which can be attributed to the well-known overestimation bias of DDPG \citep{Fujimoto2018}. The Q-value then decreases and converges, with noticeably faster stabilization under RT than under purely model-free learning. This improvement results from the transitions collected when the model-based policy is deployed in the real environment: these transitions populate the replay buffer with higher-value state--action pairs and therefore improve critic training, which in turn improves the model-free policy update in equation~\eqref{eq_policy_grad}. The model-based branch thus accelerates the model-free one, while the model-free branch enriches the data available for model assimilation. This mutual reinforcement illustrates the symbiotic nature of the RT framework.

\subsection{Policy training with online twinning and model biases}\label{C3}

The final test case combines the settings of the two previous scenarios. Training begins with the offline-calibrated M3 surrogate identified in Section~\ref{sec_off}, but this surrogate is calibrated on a database generated from a real environment with dynamics that differ from those of the environment used during control training. Online twinning is therefore required to compensate for the resulting model bias. Such drifts are realistic in flapping-wing drones, whose dynamics are difficult to characterize precisely owing to their small scale, and may evolve during operation because of damage, wear, or payload variations.

Two types of mismatch are considered relative to the original trajectory database:  
(1) a 33\% increase in body mass (referred to as the \emph{mass bias} case), and  
(2) a 3 mm longitudinal shift of the center of pressure (referred to as the \emph{cop bias} case).

Both the model-free and model-based policies are initialized randomly, and the hyperparameters are the same as those listed in Table~\ref{tab_trainParameters}. The assimilation buffer is updated using the maximum-variance strategy.

\begin{figure*}[ht]
    \begin{subfigure}{0.48\textwidth}
        \includegraphics[width=\textwidth]{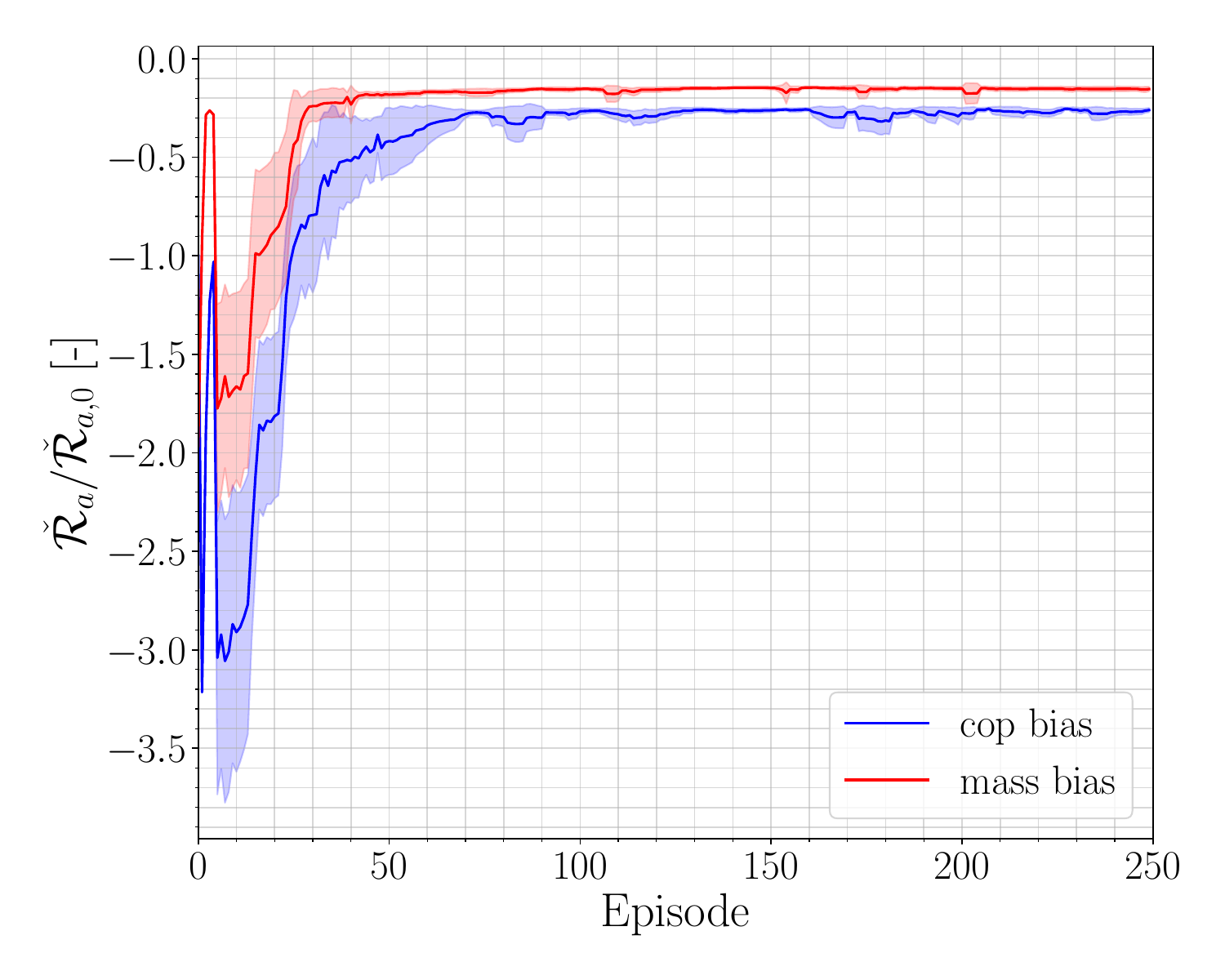}
        \caption{}
        \label{fig:case3_learningControl_a}
    \end{subfigure}
    \begin{subfigure}{0.48\textwidth}
        \includegraphics[width=\textwidth]{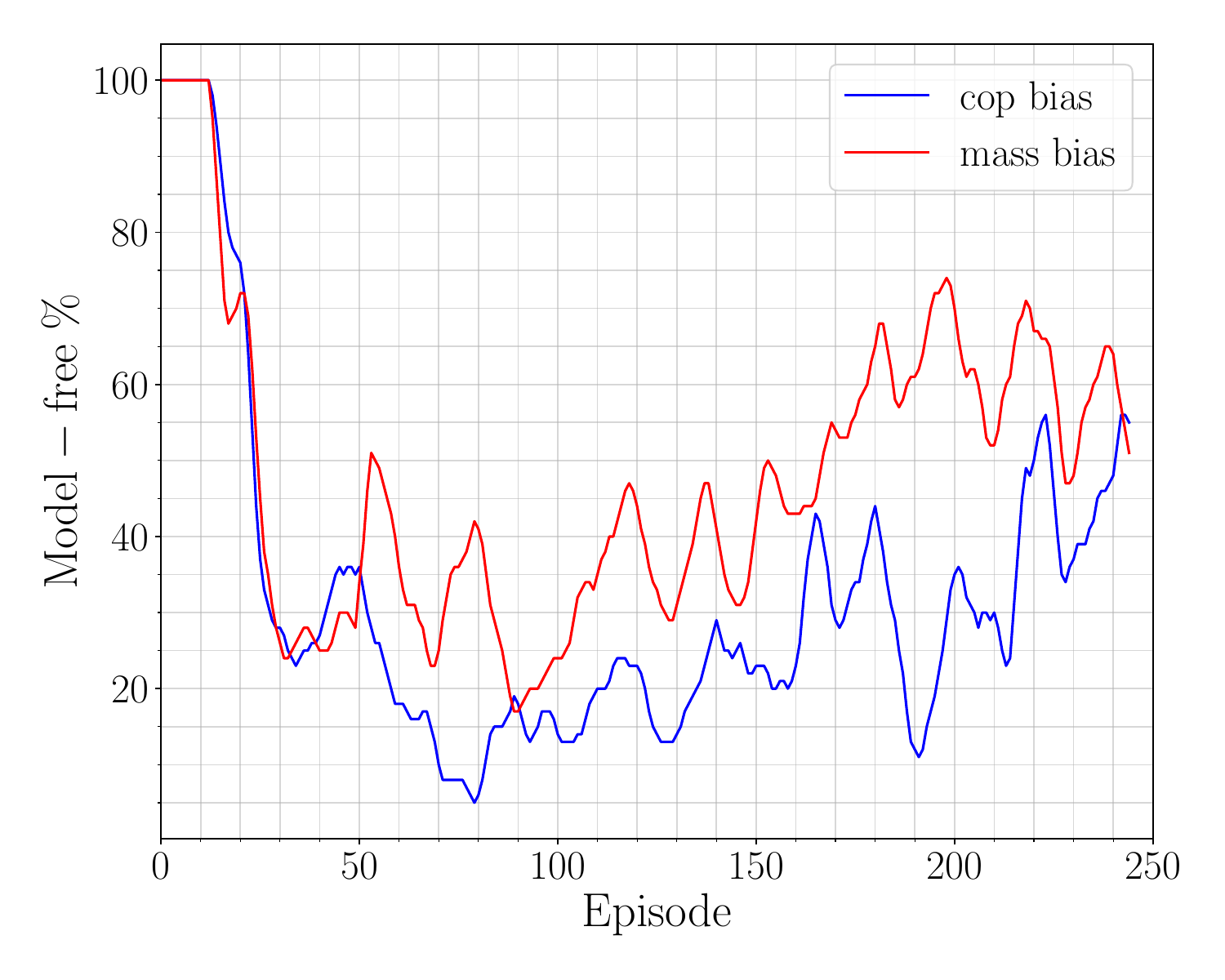}
        \caption{}
        \label{fig:case3_learningControl_b1}
    \end{subfigure}
    \begin{subfigure}{0.48\textwidth}
        \includegraphics[width=\textwidth]{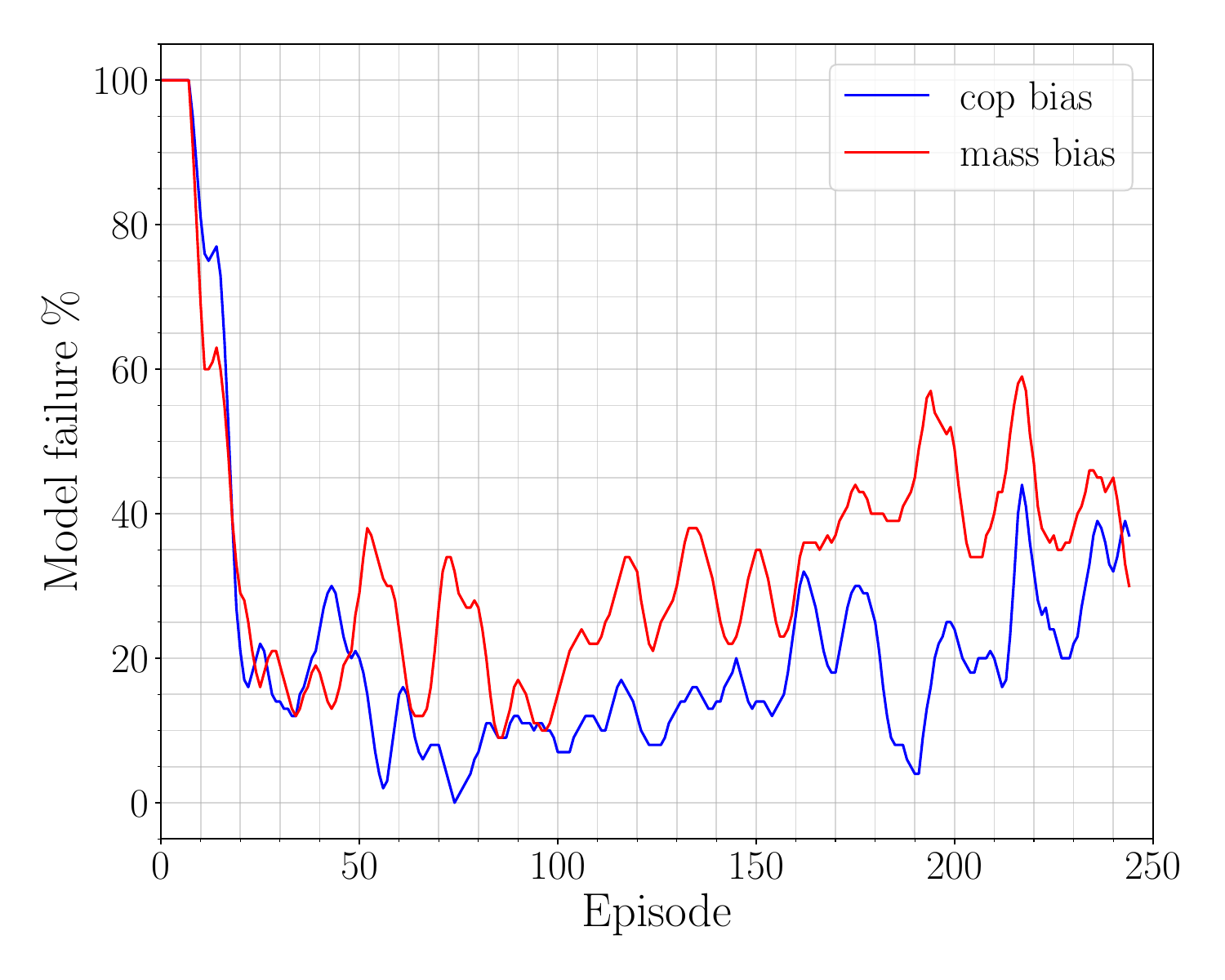}
        \caption{}
        \label{fig:case3_learningControl_b2}
    \end{subfigure}
    \begin{subfigure}{0.48\textwidth}
        \includegraphics[width=\textwidth]{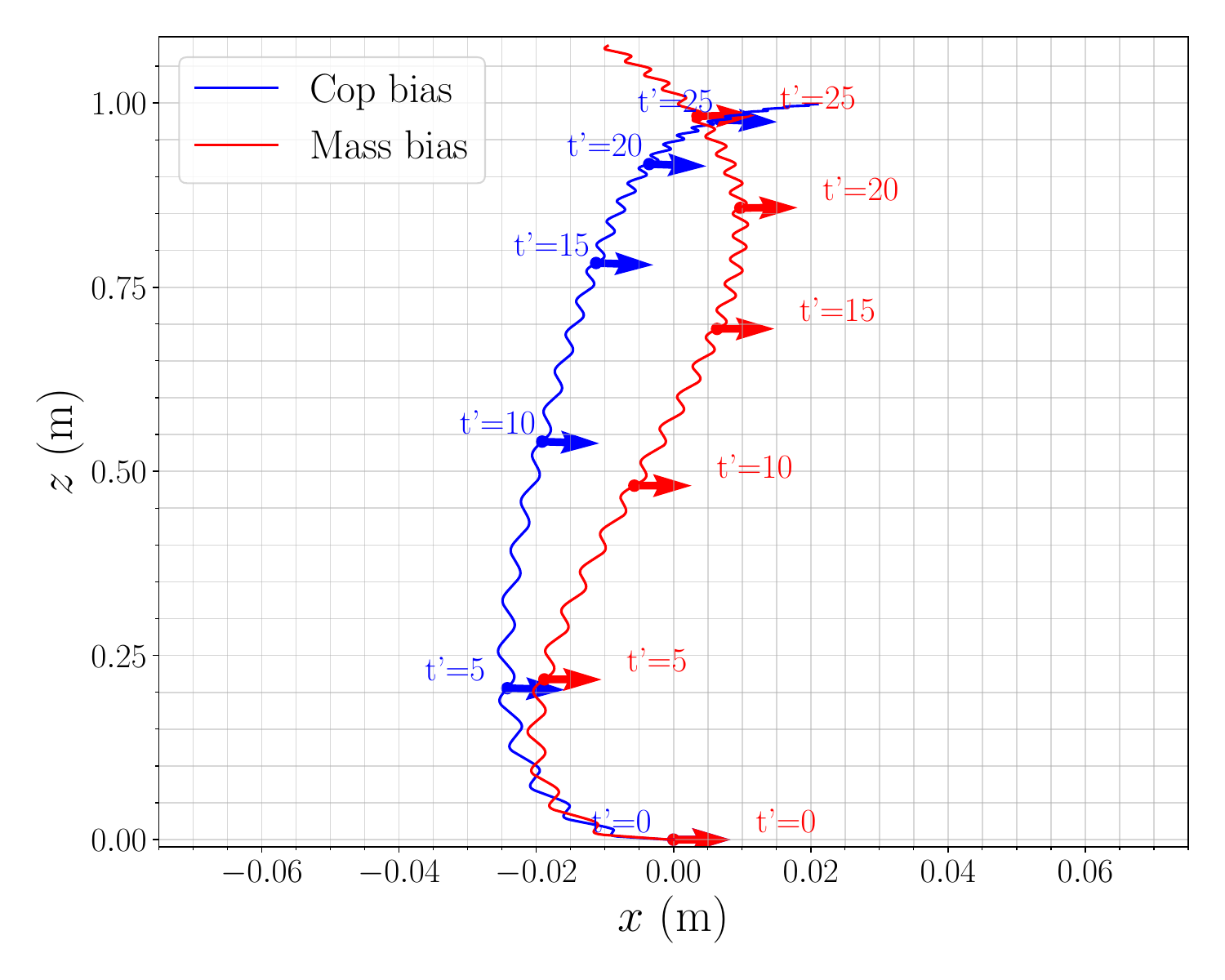}
        \caption{}
        \label{fig:case3_learningControl_d}
    \end{subfigure}
    \caption{(a) Cumulative rewards for the RT approach using the maximum-variance assimilation buffer, starting from a virtual environment calibrated offline with estimation biases in mass and centre of pressure (cop). (b) and (c) show the model-free and model-failure percentages during training for the two cases. (d) shows the resulting $x-z$ trajectories of the drone during vertical ascent.}
    \label{fig:case3_learningControl}
\end{figure*}

As in Section~\ref{C2}, Figure~\ref{fig:case3_learningControl_a} reports the mean cumulative reward and 95\% confidence interval over $N_s$ runs, while Figures~\ref{fig:case3_learningControl_b1} and \ref{fig:case3_learningControl_b2} show the fraction of runs using the model-free policy and the fraction of runs for which the surrogate is considered inconsistent with the real environment according to equation~\eqref{eq_MET}. At the start of training, the model is not yet trustworthy and the model-free branch therefore dominates. As real-environment trajectories are collected, the assimilation buffer is progressively enriched and the closure law adapts to the biased dynamics. Confidence in the surrogate is then gradually recovered, as evidenced by the decrease in model-free usage and the improved consistency between real and virtual environments. Once this occurs, the cumulative reward increases rapidly and converges with low variance. The contribution of the model-based branch is visible in the sharp reward increase observed during the early part of training.

Figure~\ref{fig:case3_learningControl_d} shows the mean $x$--$z$ trajectories obtained after training for the mass-bias and cop-bias cases. Arrows indicate the drone heading every five flapping cycles. In both cases, the drone ascends vertically and stabilizes near the target while maintaining a nearly horizontal attitude ($\theta \approx 0^\circ$), with only small longitudinal oscillations. In the mass-bias case, a slight overshoot is observed near the final vertical position. This could likely be reduced by a modest retuning of the reward weights in equation~\eqref{eq_rk}.

\section{Conclusion and perspectives}\label{sec_conclusion}

This work has presented an extended formulation of the reinforcement twinning (RT) algorithm \citep{Schena2023} and applied it to the control of flapping-wing drones. The RT algorithm combines a model-free policy trained via actor-critic reinforcement learning and a model-based policy optimized through adjoint-based data assimilation, with both agents interacting via a dynamic policy referee and supporting each other's learning.

Compared to the original formulation, the RT framework has been significantly enhanced through the introduction of a trust-based policy selection mechanism, improved collaboration strategies, and full online model identification. A cycle-averaged surrogate model incorporating second-order terms was trained using adjoint-based data assimilation, providing a computationally efficient and adaptive virtual environment.

The improved RT algorithm was evaluated across three control scenarios, varying in prior knowledge and model bias. When the virtual environment was pre-calibrated, RT demonstrated significantly higher sample efficiency than standard reinforcement learning, with the model-based agent driving early training. In fully online settings, the RT framework enabled rapid model adaptation and consistently outperformed both standalone model-free and model-based approaches. Even in the presence of substantial model bias, RT successfully corrected the surrogate model and maintained efficient policy learning within a limited number of episodes. These results confirm the benefits of the hybrid architecture and, more broadly, demonstrate the potential of RT as a general-purpose learning framework for real-time control in dynamically evolving environments.

Future work will assess the accuracy and stability of the ascending flights using CFD simulations. To test diverse flight scenarios, the simulation framework will also be further extended to include the full 3D flapping dynamics, incorporating coupled wing-body effects in the real environment. These additions may reduce the accuracy of the current surrogate models during agile maneuvers, prompting ongoing exploration of enhanced closures, local model interpolation, or time-varying formulations. A key direction will be to further integrate the two learning modes, allowing the model-based policy to more directly exploit real-environment experience gathered by the model-free agent.

\section*{Acknowledgements}
Romain Poletti and Lorenzo Schena were supported by Fonds Weten-
schappelijk Onderzoek (FWO), Project No. 1SD7823N and 1S67925N. The work was also partially supported by the European Research Council (ERC) under the European Union’s Horizon Europe research and innovation programme, through a Starting Grant awarded to M. A. Mendez (grant agreement No. 101165479 — RE-TWIST). The views and opinions expressed are those of the authors only and do not necessarily reflect those of the European Union or the European Research Council. Neither the European Union nor the granting authority can be held responsible for them.









\bibliographystyle{cas-model2-names}

\bibliography{Biblio}




\end{document}